# Prediction and Forecast of Short-Term Drought Impacts Using Machine Learning to Support Mitigation and Adaptation Efforts

Hatim M. E. Geli[1,2,*], Islam Omar[3,*], Mona Y. Elshinawy[4], David W. DuBios[5], Lara Prehodko[6], Kelly H Smith[7], Abdel-Hameed A. Badawy[3]

[1] Department of Animal and Range Science, New Mexico State University, Las Cruces, NM 88003, USA
[2] New Mexico Water Resources Research Institute, New Mexico State University, Las Cruces, NM 88003, USA
[3] Klipsch School of Electrical and Computer Engineering, New Mexico State University, Las Cruces, NM 88003, USA
[4] Department of Engineering Technology and Surveying Engineering, New Mexico State University, Las Cruces, NM 88003, USA
[5] Department of Plant and Environmental Sciences, New Mexico State University, Las Cruces, NM 88003, USA
[6] College of Agriculture, Consumer, and Environmental Sciences, New Mexico State University, Las Cruces, NM 88003, USA
[7] National Drought Mitigation Center, School of Natural Resources, University of Nebraska-Lincoln, Lincoln, NE 68583, USA

[*] co-first authors

## Abstract

Drought is a complex natural hazard that affects ecological and human systems, often resulting in substantial environmental and economic losses. Recent increases in drought severity, frequency, and duration underscore the need for effective monitoring and mitigation strategies. Predicting drought impacts—rather than drought conditions alone—offers opportunities to support early warning systems and proactive decision-making.

This study applies machine learning techniques to link drought indices with historical drought impact records (2005–2024) to generate short-term impact forecasts. By addressing key conceptual and data-driven challenges, the study aims to improve the predictability of drought impacts at actionable lead times. The Drought Severity and Coverage Index (DSCI) and the Evaporative Stress Index (ESI) were combined with impact data from the Drought Impact Reporter (DIR) to model and forecast weekly drought impacts.

Results indicate that Fire and Relief impacts were predicted with the highest accuracy, followed by Agriculture and Water, while forecasts for Plants and Society impacts showed greater variability. County- and state-level forecasts for New Mexico were produced using an eXtreme Gradient Boosting (XGBoost) model that incorporated both DSCI and ESI. The model successfully generated forecasts up to eight weeks in advance using the preceding eight weeks of data for most impact categories.

This work supports the development of an Ecological Drought Information Communication System (EcoDri) for New Mexico and demonstrates the potential for broader application in similar drought-prone regions. The findings can aid stakeholders, land managers, and decision-makers in developing and implementing more effective drought mitigation and adaptation strategies.

# 1. Introduction

Drought is a complex natural hazard that develops gradually over time, posing serious threats to both human and natural ecosystems (Wilhite *et al* 2007). Growing concern surrounding drought events stems from their widespread and often severe impacts, which include loss of life, economic damage, and long-term environmental degradation. Since 1980, the United States has experienced 31 major drought events, resulting in an estimated $362 billion in damages, making drought the third costliest disaster type and accounting for approximately 13% of total disaster-related costs (NOAA NCEI, 2025). Unlike other natural hazards, drought lacks visually apparent markers, yet its consequences can be profound. Therefore, effective drought preparedness and mitigation require not only monitoring current conditions but also anticipating potential impacts (UNDRR, 2021; WMO, 2025). Predicting these impacts is essential for developing early warning systems, such as NOAA's Drought Early Warning Systems (NOAA NIDIS 2025), and for empowering resource managers, government agencies, and other stakeholders to take proactive and informed action.

Drought is conceptually defined as a prolonged precipitation deficit (Wilhite, 2000) and can manifest in multiple forms, including meteorological, hydrological, agricultural, and socioeconomic droughts, as well as recently recognized flash and ecological droughts (Crausbay *et al.,* 2017; Wilhite & Glantz, 1985; Otkin *et al.,* 2018). Each type affects physical and social systems differently, often producing compounding impacts such as water shortages, wildfires, crops and livestock losses, vegetation decline, and livelihood disruptions. Linking drought types to their impacts is critical for effective management but remains challenging (Bachmair *et al.,* 2015; 2016; 2017; Sandholt *et al.,* 2002; Zhang *et al.,* 2023). Drought monitoring relies on indicators such as precipitation and temperature, summarized through indices like the Evaporative Stress Index (ESI), which describe the onset, duration, severity, and extent (WMO, GWP 2016). These indices can be associated with text-based impact reports, providing opportunities for ground-truthing and verification. However, conceptual challenges persist: First, no single index can capture all types of droughts. Second, the effectiveness of drought indices in representing specific drought conditions relies heavily on a convergence of evidence since drought lacks direct observable indicators (Van Loon *et al.,* 2016a; 2016b; Noel *et al.,* 2020). Third, impacts often result from cascading effects. Recent studies combine multiple indices to predict impacts (Zhang *et al.*, 2023; Noel *et al.*, 2020; Sutanto *et al.*, 2019), but gaps remain. This study approaches these challenges conceptually by examining how drought impacts are represented and how well existing drought indices capture the temporal and spatial dimensions of those impacts.

To address key conceptual challenges and establish functional relationships between drought indices and impacts, this analysis focuses on two main aspects: (1) how the temporal and spatial characteristics of drought impacts are numerically represented, and (2) how existing drought indices reflect those impacts. Developing meaningful relationships between drought indices and their associated impacts requires well-represented, high-quality drought impact data. However, limited and imbalanced datasets have posed significant obstacles, particularly in tackling the second and third challenges outlined earlier. Some studies have attempted to mitigate this issue by aggregating data. For example, Sutanto et al. (2019) and Bachmair et al. (2017) aggregated data across broad geographic regions and impact categories within the European Drought Impact Report Inventory (EDII). Similarly, Noel et al. (2020) and Zhang et al. (2023) aggregated U.S. Drought Impact Reporter (DIR) data from counties to states or by time and impact category at the national level. Although aggregation and cost-sensitive learning approaches offer partial solutions, they do not fully address the underlying imbalance (Zhang *et al.,* 2023). Moreover, aggregation can obscure critical information: temporal aggregation limits fine-scale (e.g., weekly) analysis; spatial aggregation reduces insight into localized impacts and spatial dependencies; and categorical aggregation masks relationships between specific impacts and drought events. To overcome these limitations, this study employs an advanced data augmentation technique known as Synthetic Minority Over-sampling (SMOTE), available through the Imbalanced-Learn library. Data augmentation was essential in this analysis, given the goal of obtaining higher resolution predictions (*i.e.*, weekly and county-level), which exacerbates the imbalance in drought impact datasets.

Moreover, temporal aggregation presents a challenge in capturing the rapid development of certain physical processes that lead to drought events and their associated impacts. This limitation can hinder the creation of robust models that effectively link drought indices to observed impacts. While recent studies have typically employed a monthly time scale—the finest resolution used so far—some droughts, such as the 2012 flash drought, have been observed to develop within just a few days to weeks, with impacts reported over similarly short periods (Leeper *et al* 2022, Otkin *et al* 2018). Over the past decade, droughts have become increasingly intense, developing quickly (rapid intensification) and persisting longer (Otkin *et al* 2016, 2018). Consequently, relying solely on monthly data risks missing key features of flash droughts, such as sudden crop losses or rapid vegetation decline. Using a weekly time step improves representation of these dynamic processes and allows for the inclusion of both the presence and absence of reported impacts, acknowledging that drought conditions may exist without immediate observable effects. In this analysis, drought impact modeling was conducted at a weekly temporal resolution and county-level spatial scale, allowing for consideration of spatial correlations across neighboring counties – a crucial factor, given that droughts often extend beyond administrative boundaries.

Additionally, identifying a drought index capable of representing various types of droughts and capturing their multidimensional impacts remains a significant challenge. Because drought impacts stem from diverse physical processes, relying on a single indicator is often insufficient. Previous studies have employed drought indices based on single or multiple indicators, such as the Standardized Precipitation Index (SPI), Standardized Temperature Index (STI), and Standardized Precipitation Evapotranspiration Index (SPEI) (Zhang *et al.*, 2023; Bachmair *et al.*, 2017; Sutanto *et al.*, 2019). This analysis leverages the unique characteristics of the U.S. Drought Monitor (USDM) (Svoboda *et al.*, 2002; USDM 2019), which offers a quasi-quantitative and quasi-observational depiction of drought events. The USDM is quasi-observational because it integrates numerical data with on-the-ground assessments from approximately 450 drought experts, including State Climatologists, thereby intrinsically incorporating drought impacts. Although the USDM provides categorical classification of drought conditions, it has been quantitatively represented by the Drought Severity Coverage Index (DSCI) (USDM - DSCI 2021), which enables meaningful analysis of the relationship between drought events and their impacts (Noel *et al.*, 2020; Leeper *et al.*, 2022; Johnson *et al.*, 2020). In this study, the DSCI was utilized and further complemented by the ESI, which is particularly effective in detecting rapidly intensifying drought conditions (Otkin *et al* 2018). Given the nonlinear nature of these relationships, advanced machine learning techniques, such as the eXtreme Gradient Boosting (XGBoost) algorithm (Chen & Guestrin, 2016), were employed.

This study focused on evaluating drought impacts across New Mexico, a state where drought is a recurring climatic phenomenon. The World Resources Institute (WRI) classified New Mexico as facing "Extremely High" water quantity risk (WRI 2020). With an annual average precipitation of 37.1 cm (14.6 inches) from 1895 to 2020, its ecosystems have adapted to semi-arid and arid climate conditions (Abatzoglou *et al.*, 2017; WRCC 2025). However, ongoing climate changes are increasing the state's vulnerability to drought-related impacts. Notably, since January 2000,

New Mexico has experienced at least a "Moderate Drought (D1)" for 1,096 weeks – accounting for 99.5% of the time—more than any other US state (Zhang *et al.,* 2023). These factors underscore the critical need for accurate predictions of drought impacts.

The goal of this analysis is to predict drought impacts and forecast their occurrence with a reasonable lead time, thereby helping individual stakeholders, land managers, and decision-makers develop and implement mitigation and adaptation plans. To achieve this goal the analysis addressed the following objectives: 1) develop drought impact prediction models using the DSCI and ESI at the state and county levels and at a weekly time scale, 2) evaluate these models to identify predictions with reasonable lead-time for the different drought impact categories, 3) evaluate the effects of spatial correlation of drought conditions and impacts from neighboring counties, and 4) highlight limitations of current drought impacts data and how ML was used to address them. This work is conducted within the context of developing the Ecological Drought Information Communication System (EcoDri) (Geli, 2025), and the modeling framework developed here can be generalized across the United States and adapted for other drought-prone regions.

## 2. Data

This section describes the study area, drought impacts, and monitoring tools, as well as data pre-processing, data expansion, and machine learning algorithms used to predict drought impacts, along with evaluation metrics.

### 2.1. Study Area

This study was conducted over New Mexico (NM) in the southwestern US (Figure 1). New Mexico's regional climate is arid to semi-arid, with a mean annual precipitation ranging from 250 mm in the southern Chihuahuan Deserts to 500 mm in the northern mountains. The state's mean annual air temperature is about 12°C. The natural vegetation cover is mostly grassland, shrubland, and forest (Gedefaw *et al.,* 2021). NM's land is predominantly used for grazing and crop production practices that rely on surface and groundwater; mining and extraction of fossil fuels, including oil, gas, and coal; and tourism (Yadav *et al.,* 2021). The combination of land use and land cover; increased temperature; variable precipitation; and increased pace of climate change during the past 3 – 4 decades have created ideal conditions for extreme climate events such as droughts. These drought events have resulted in cascading impacts on social and ecological systems across New Mexico (Johnson *et al.,* 2020) (Figure 2). This study evaluated drought impacts that occurred during the period between January 2005 and December 2024.

### 2.2. Drought Indices
#### 2.2.1. Drought Severity and Coverage Index (DSCI)

The U.S. Drought Monitor (USDM) is a unique drought monitoring tool developed by the National Drought Mitigation Center (NDMC) to provide a weekly map-based assessment of drought severity across the US. The USDM is based on an assessment of drought conditions using a wide range of inputs representing the hydrologic cycle, including real-time climate, water, and remotely sensed data. A defining feature of the USDM is its reliance on continuous, on-the-ground feedback from hundreds of drought experts nationwide, which enhances the accuracy and contextual relevance of its weekly assessments (Svoboda *et al.,* 2002). The Drought Severity and Coverage Index (DSCI) is an experimental method developed to convert categorical USDM drought levels into a single, continuous, aggregated value for a specified area, thereby determining special coverage and intensity more effectively (Akyuz, 2017; Smith *et al.,* 2020). The DSCI is calculated by assigning a weighted average of 1 through 5 to each USDM category D0 – D4. This weight is multiplied by the categorical percent area for the drought category, and these totals are summed together. This results in a DSCI value that has a continuous scale of 0–500. Two advantages can result from converting the percent area in each USDM drought category into the DSCI. First, it provides a single numerical value describing current drought extent and intensity. Second, it allows users to quantify drought over time, more meaningfully than separate percentages of area coverage for each category. Although the USDM effectively depicts the spatial distribution of drought through its real-time maps, it does not provide an inherently straightforward way to analyze drought conditions over time or compare them directly with other drought indices (Johnson *et al.,* 2020).

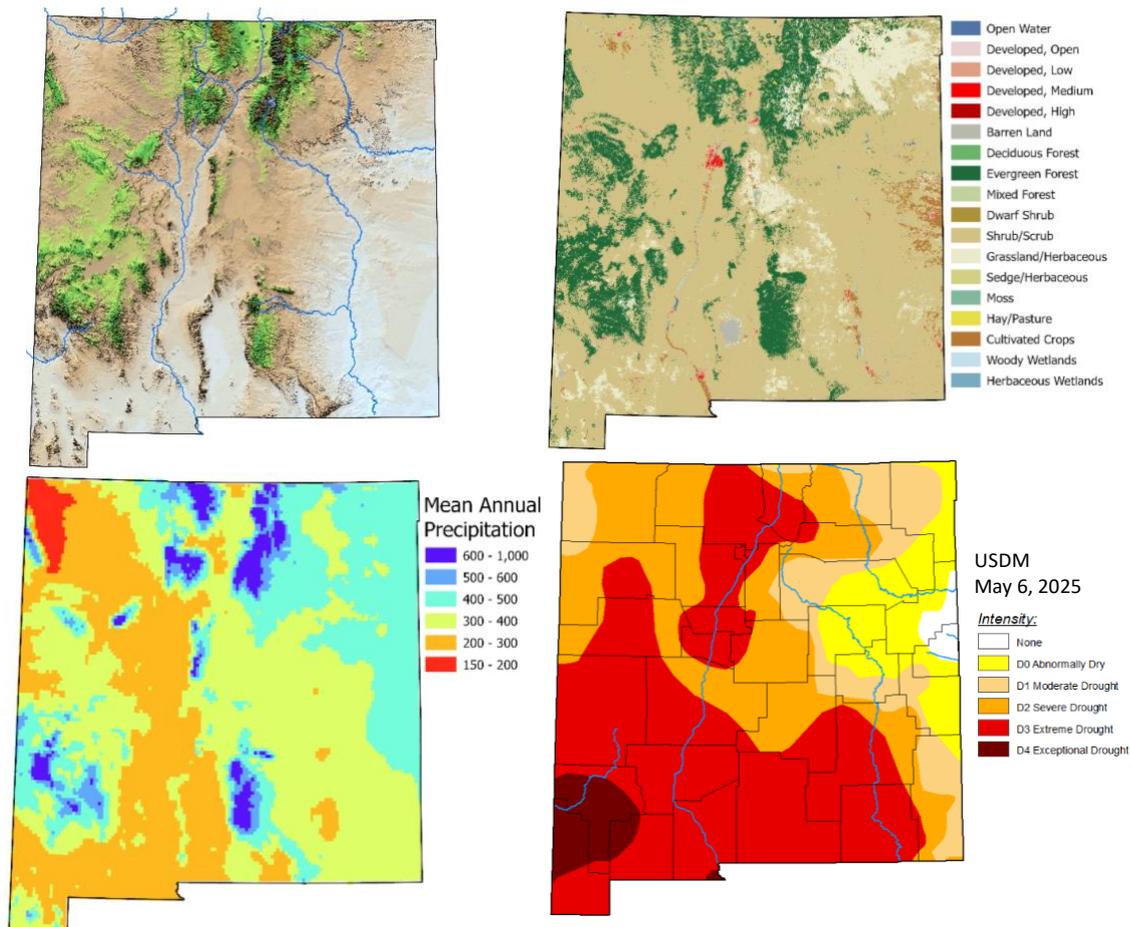

Figure 1: New Mexico's topography and major river [modified from ref], land cover based on the National Land Cover Data (NLCD) 2023 (Jin *et al* 2023), mean annual precipitation based for the period between 2000 – 2025 based on PRISM (PRISM 2025), drought condition for the week ending on May 6, 2025, based on the USDM (USDM 2019).

### 2.2.2. Evaporative Stress Index

The Evaporative Stress Index (ESI) (Anderson *et al.,* 2007a, 2007b) is a standardized measure of the ratio of actual-to-potential evapotranspiration (fPET = ET/PET), where ET and PET are instantaneous or daily estimates derived using the ALEXI model (Anderson *et al.,* 2011). Normalization by PET serves to minimize variability in ET due to seasonal variations in available energy and vegetation cover, refining focus on the soil moisture signal. The ESI has demonstrated its ability to serve as a proxy for surface soil moisture conditions, which are crucial for monitoring agricultural and vegetation water stress (Anderson *et al.,* 2011; Otkin *et al.,* 2015; 2016). The currently operational ESI product that is available for North America is based on Moderate Resolution Imaging Spectroradiometer (MODIS) satellite data at a 4-km scale (SERVIR GLOBAL 2025). ESI has been used to identify flash drought and agricultural drought events (Otkin *et al.,* 2016; Raghav and Kumar, 2024).

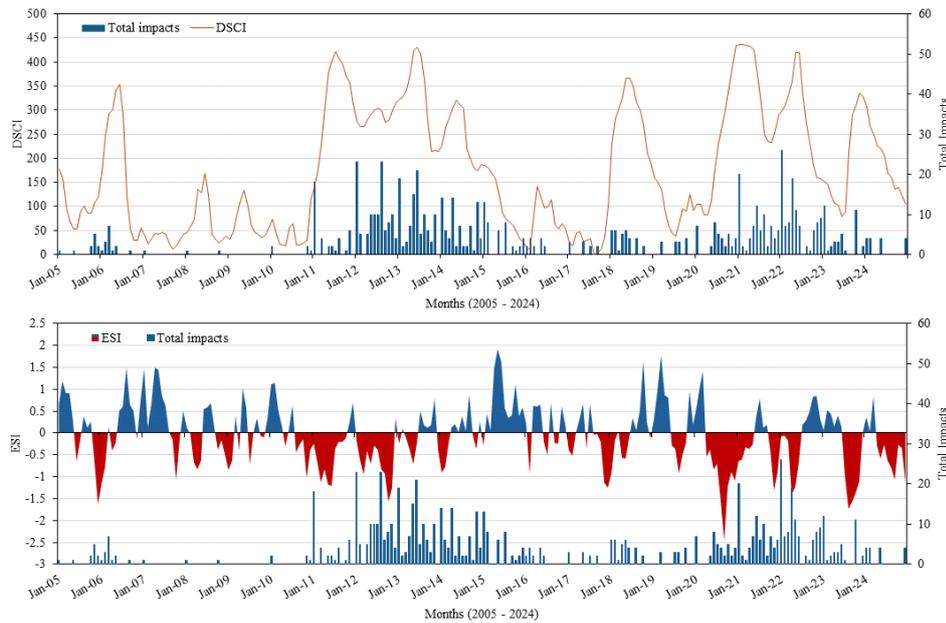

Figure 2: New Mexico's drought impacts (DIR 2025), Drought Severity Coverage Index (DSCI) (USDM 2020), and Evaporative Stress Index (ESI) (SERVIR GLOBAL 2025) for the period between January 2005 and December 2024 (modified from (Johnson *et al.,* 2020)).

Table 1: Descriptions of drought categories as defined by the National Drought Mitigation Center (DIR Dashboard 2025).

| Impact Category (short name) | Description |
|---|---|
| Agriculture | Drought affecting agriculture, farming, aquaculture, horticulture, forestry, or ranching. |
| Water Supply & Quality (*Water*) | Drought affecting water supply and water quality, and related activities such as restrictions, allocation, and mitigation. |
| Business & Industry (*Business*) | Drought affecting non-agriculture and non-tourism businesses, such as a reduction or loss of demand for goods or services, reduction in employment and other economic impacts. |
| Energy | Drought affecting power production, rates, and revenue. |
| Fire | Drought affecting forest, range, rural, or urban fires, fire danger, and burning restrictions. |
| Plants & Wildlife (*Plants*) | Drought affecting unmanaged plants and wildlife, both aquatic and terrestrial. |
| Relief, Response & Restrictions (*Relief*) | Drought affecting disaster declarations, aid programs, requests for disaster declaration or aid, water restrictions, or fire restrictions. |
| Society & Public Health (*Society*) | Drought affecting human, public, and social health, including health-related problems. |
| Tourism & Recreation (*Tourism*) | Drought affecting recreational activities and tourism. |

## 2.3. Drought Impacts

Unlike other natural hazards, the impacts of drought are often difficult to detect until they begin to materialize. In many cases, impacts may continue even after drought indicators suggest that conditions have improved, and there is no universally accepted definition for when a drought officially ends. Because of the complexity of possible impacts, drought impact collection and drought monitoring have always been difficult. This lack of impact information in decision-making leads to an inability to capture the full score of drought situations (Redmond 2002). To address these challenges, the National Drought Mitigation Center (NDMC) launched the Drought Impact Reporter (DIR 2025) in 2005, creating the nation's first comprehensive, web-based archive dedicated to documenting drought impacts (Table 1). The DIR is a moderated database of events primarily drawn from news stories that document the occurrence of drought impacts. Impacts are categorized by sector (or category) and can be mapped for different categories and locations. For this study, drought impacts (Table 1) were downloaded from the DIR for the 33 counties in New Mexico (DIR Dashboard 2025).

## 3. Methods
### 3.1. Modeling Framework

To support the development of drought impact predictions and facilitate future applications, we developed a modeling framework referred to as the Drought Impacts Prediction (DIP) framework (Figure 3). The DIP framework consists of four sequential stages: (1) data integration, (2) data augmentation, (3) ML models development, and (4) forecast evaluation. A key feature of the DIP framework is the use of a loop mechanism, which systematically selects one county at a time, excluding its data from the training set, and trains the model on the data from the remaining counties. This process was repeated for each county to ensure robust and generalizable model performance. The modeling process was executed iteratively across varying durations of antecedent drought conditions. We hypothesized that model performance would vary depending on both the duration and severity of drought exposure. The DIP framework considered drought conditions extending up to eight weeks prior to the observed impacts. Accordingly, separate ML models were trained for each impact type using multiple temporal configurations, including one to eight weeks, two to eight weeks, and other combinations, to capture the temporal dynamics of drought-related effects.

### 3.2. Data Pre-processing and Integration

To map impacts to drought indices, a preliminary analysis was conducted to develop a consistent climatology of drought indices, along with associated reported impacts, thereby facilitating better predictions of impacts when triggered by drought-specific events. Developing this climatology is key to obtaining a model with high skill. Drought impacts are reported when they are observed. In some cases, however, drought impacts may be reported during the same week of the drought index or afterward. That means impacts can appear and be reported after a prolonged period of drought conditions, but not immediately. It was assumed in this analysis that impacts occurred during the same week of the reported drought indices. A binary setting was considered for the impact data, with "one" and "zero" to indicate impact and no impact.

The data was organized in rows (representing temporal instances) and columns (representing features). Visualizing the timeseries (Figure 4) revealed two crucial data challenges:
1. **Limited Features:** A limited number of features on which to train the ML model.
2. **Class Imbalance:** A severe imbalance between the two binary classes. Instances representing an impact (class "one") were significantly underrepresented, while non-impact instances (class "zero") dominated the dataset.

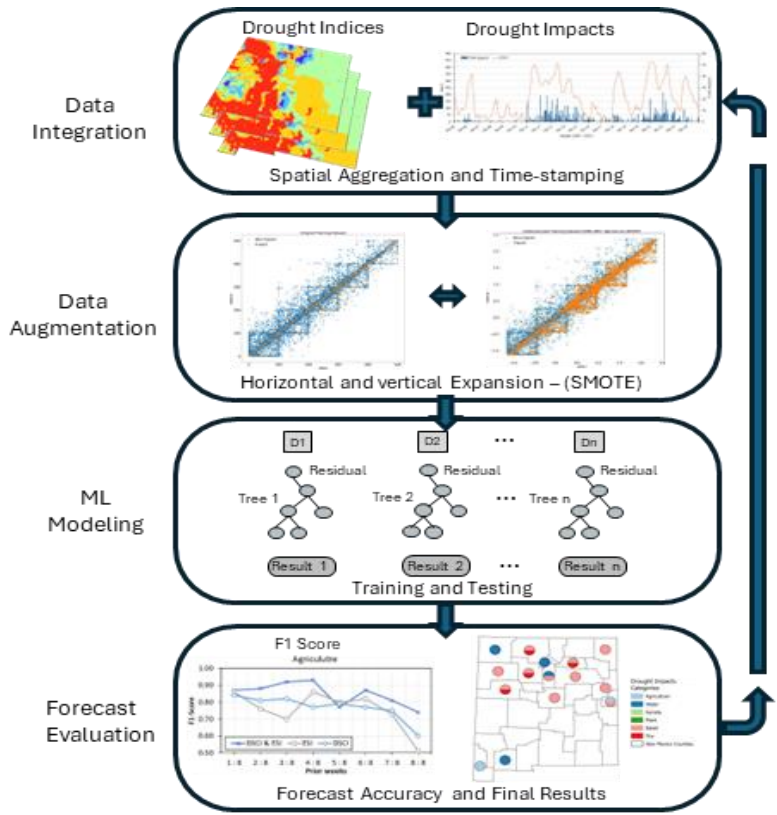

Figure 3: A depiction of the Drought Impacts Prediction (DIP) framework.

These challenges necessitated both horizontal (feature) and vertical (instance) data expansion, as described in the next section. The drought indices (DSCI, ESI) and binary impacts over the 33 counties were organized in a matrix format to reflect a weekly combination. The impact columns exclusively contain binary values (1 for presence, 0 for absence). A preliminary analysis of the dataset (Figure 4) indicated quasi-decadal oscillation and raised three issues:
1. Impacts could occur under both high and low DSCI values.
2. A clear trend emerged in the DSCI values over the years.
3. DSCI values did not exhibit substantial variations within short timeframes (*i.e.*, weekly).

These observations informed the consideration of two modeling approaches: Time Series Classification (*e.g.*, Long Short-Term Memory, LSTM) and Supervised ML Models (*e.g.*, Random Forest, XGBoost). The latter requires considerable data, reinforcing the need to integrate data from neighboring counties and apply data augmentation. This approach aligns with our objective of forecasting impacts with sufficient lead time.

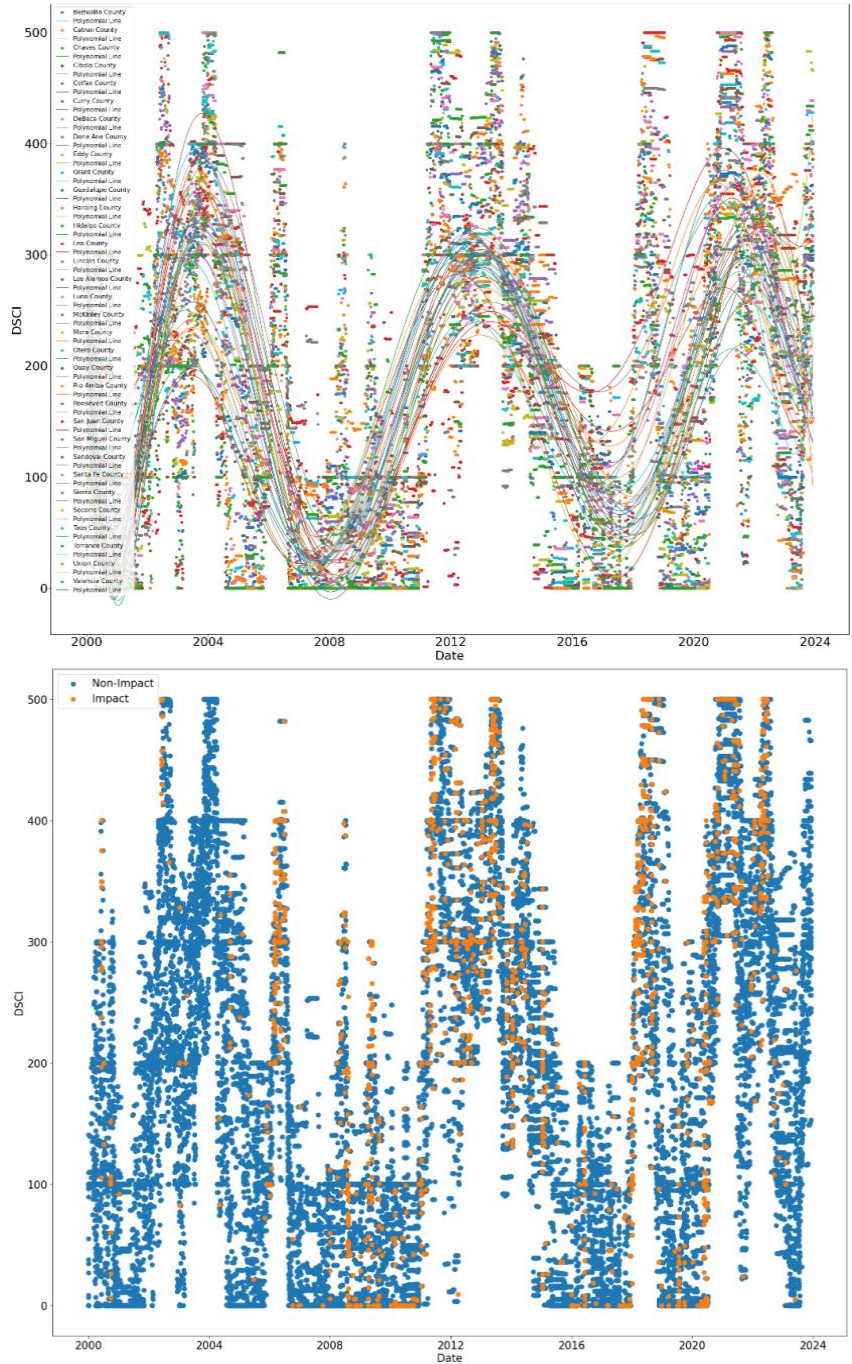

Figure 4: The top panel shows the weekly Drought Severity Coverage Index (DSCI) for each county in New Mexico, overlaid with fitted seasonality. The bottom panel displays the DSCI, along with the reported total impacts in orange and blue, indicating positive and negative impacts, respectively.

### 3.3. Data Augmentation

Drought impact data is known for its limited availability. This limitation was addressed using the following ML tools and techniques.

#### 3.3.1. Horizontal Expansion (Neighboring Counties)

We augmented the dataset features by introducing data from neighboring counties, specifically including the previous eight weeks of their drought indices and corresponding impacts. This extension broadened the dataset horizontally. We made this choice with the understanding that an impact in a neighboring county heightened the likelihood of a similar impact in the target county. However, this posed a new challenge: not all counties have an equal number of neighbors, resulting in an imbalance in feature columns and missing fields for neighbors' data that could not be filled (as the data did not exist). This challenge heavily influenced our selection of ML models.

### 3.3.2. Vertical Expansion (Oversampling and Undersampling)

To address the severe class imbalance in the target "impacts" columns, we employed a two-step process using the **Imbalanced-Learn** library (Lemaître *et al* 2017).

1. **Oversampling (Borderline SMOTE):** We first tested standard **Synthetic Minority Oversampling (SMOTE),** but observed that it did not adequately consider regions of higher data density, which was confusing for the models (Figure 5b). We ultimately opted to use **Border Line SMOTE**, a variant that generates new synthetic samples specifically within these high-density regions (Figure 5c). This approach was ideal for our data, as it helped the model account for the fact that impacts could occur even at low DSCI values.
2. **Undersampling (ENN):** After oversampling the minority (impact) class, we "cleaned" the dataset by under-sampling the majority (non-impact) class using **EditedNearestNeighbours (ENN)**. This process removes synthetically generated instances that could lead to model divergence, resulting in a final, smoothed training dataset (Figure 5d).

All data was normalized before applying oversampling techniques, following machine learning standards.

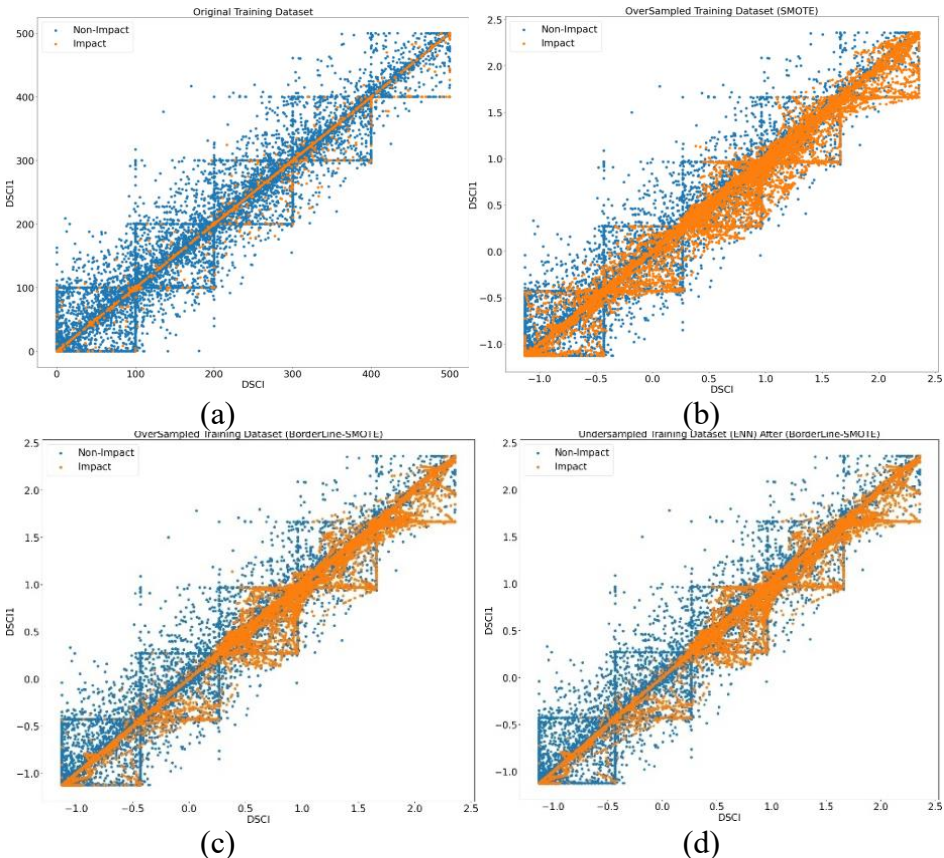

Figure 5: (a) The original data with the main DSCI on the X-axis and DSCI-1 on the Y-axis. (b) The oversampled data using SMOTE with DSCI on the X-axis and DSCI-1 on the Y-axis. (c) The oversampled data using Border Line SMOTE with DSCI on the X-axis and DSCI-1 on the Y-axis. (d) The under-sampled data generated using ENN is plotted against the oversampled data, which was generated using Border Line SMOTE as input, with DSCI on the X-axis and DSCI-1 on the Y-axis.

### 3.4. Machine Learning Models

Three ML models were evaluated, including Random Forest (RF), Long Short-Term Memory (LSTM), and eXtreme Gradient Boosting (XGBoost). These models have been tested in similar applications and have demonstrated potential for effective performance (Bachmair *et al.,* 2017; Zhang *et al.,* 2023; Sutanto *et al.,* 2019). RF and XGBoost were chosen as powerful, tree-based ensemble methods suitable for tabular data. LSTM was selected as a deep learning model specifically designed to capture long-term temporal dependencies in time series data. Our LSTM model architecture consisted of three LSTM layers, each with 100 units. The first two layers used return_sequences=True, while the third did not. Dropout (rate of 0.3) was applied after each LSTM layer to reduce overfitting. A Dense layer (50 units, ReLU activation) followed the LSTM layers, and the output layer used 1 unit with a sigmoid activation function for binary classification. The model was compiled with the Adam optimizer and binary cross-entropy loss.

### 3.5. Model Implementation and Training

The inclusion of neighbors' data required a reevaluation of the models due to the missing attributes (varying number of neighbors per county).

For models like RF, we employed Scikit-Learn's SimpleImputer algorithm to represent the absence of neighbor data with a placeholder value (-999) instead of N/A.

However, the XGBoost library (Chen and Guestrin 2016) is known for its efficiency and, critically, its native ability to handle missing values. This made it an ideal choice for our data structure, as it did not require imputation.

The final, successful workflow for constructing our XGBClassifier model for each impact type was as follows:

1. **Feature Selection:** Select feature columns based on prior weeks' data (drought indexes and impacts) for the target county and its neighboring counties.
2. **Data Split:** Split the data into 80% training and 20% testing sets.
3. **Handle Missing Data:** For RF, use SimpleImputer to fill NA values with -999. For XGBoost, this step was handled natively by the algorithm.
4. **Normalization:** Normalize the data using StandardScaler.
5. **Oversampling:** Apply BorderLine SMOTE to the training data.
6. **Undersampling:** Apply EditedNearestNeighbours to the resulting oversampled training data.

### 3.6. Predictions Evaluation and Feature Contribution

#### 3.6.1. Evaluation Metrics

The F1-score was used as the primary evaluation metric, as it is ideal for imbalanced datasets. It represents the harmonic Mean of precision and recall, balancing false positives and false negatives. A score closer to 1 indicates better performance.

$$F1 - score \ = 2 \frac{Precision \cdot Recall}{Precision + Recall} \qquad \text{(Eq. 1)}$$

where $Precision = \frac{TP}{TP + FP}$, and $Recall = \frac{TP}{TP + FN}$ (TP = true positive, FP = false positive, FN = false negative). A higher F1-score (closer to 1) indicates better performance

#### 3.6.2. Features Contribution

We evaluated the relative importance of variables (features) on the prediction quality. We refer to this as a feature contribution and use it to compare model performance. Features were grouped into four categories:
1. **DI** (Drought Index of the target county)
2. **IMPs** (Drought Impacts of the target county)
3. **NEIGH_DI** (Drought Index of neighboring counties)
4. **NEIGH_IMPs** (Drought Impacts of neighboring counties)

This analysis was conducted only for the RF and XGBoost models, as they provide built-in estimates for feature contribution.
- **RandomForest:** We used the "feature_importance" attribute, which is based on **Gini Importance** (or Mean Decrease Impurity). It estimates the decrease in impurity within a node when a feature is used to split the data, averaged across all trees.
- **XGBoost:** We used the built-in "feature_importance" function with the "**gain**" method. This calculates the squared differences between parent and child node impurities and sums them up, averaged across all trees.

## 4. Results
### 4.1. Summary Augmented Data
The original dataset contained a significant imbalance between reported impacts and non-impact instances. Table 2 presents the original count of impacts for each category, compared to the final count after applying Border Line SMOTE oversampling and ENN undersampling to balance the dataset for machine learning. The total original impacts (3,835) were substantially augmented to create a balanced training dataset.

Table 2: Summary of reported and augmented data

| Impact Category | Reported impacts (original/oversampled) | No impact (original/oversampled) | Total (original/oversampled) |
|---|---|---|---|
| Agriculture | 736 / 40481 | 40481 / 40481 | 41217 / 80962 |
| Water | 489 / 40728 | 40728 / 40728 | 41217 / 81456 |
| Fire | 715 / 40502 | 40502 / 40502 | 41217 / 81004 |
| Plants | 465 / 40752 | 40752 / 40752 | 41217 / 81504 |
| Relief | 1139 / 40078 | 40078 / 40078 | 41217 / 80156 |
| Society | 203 / 41014 | 41014 / 41014 | 41217 / 82028 |
| Tourism | 88 / 41129 | 41129 / 41129 | 41217 / 82258 |
| Total | 3835 / 284684 | 284684 / 284684 | 288519 / 569368 |

### 4.2. State-Level Predictions
All three models (LSTM, RF, and XGBoost) exhibited varying performance across drought impact categories, regardless of the drought index applied (Tables 3–5). None of the models successfully predicted impacts in the Business & Industry and Energy categories. Among them, XGBoost achieved the highest overall accuracy, with most F1-scores ranging from 0.70 to 0.94 and only a few between 0.50 and 0.69. F1-scores below 0.50 were considered unacceptable. LSTM delivered moderate performance, with scores generally ranging between 0.50 and 0.80, and occasionally exceeding 0.80. RF consistently showed the lowest performance.

XGBoost reliably predicted all impact categories across all prior-week configurations, except for the 5:8 prior-week scenario. Accuracy improved when recent data from the past weeks were used and when DSCI and ESI were combined. LSTM produced reasonable predictions for only two

categories—Relief and Fire—across all prior weeks when using ESI (except for 3:8 weeks), and for Water, Relief, and Fire when using DSCI. When combining ESI and DSCI, LSTM again predicted Relief and Fire, but performance gains were minimal. RF was limited to predicting Fire (except week 8:8) and Relief (except weeks 6–8) when using ESI; Relief and Fire when using DSCI; and similarly, Relief and Fire (except weeks 7:8) when both indices were combined.

Using ESI and DSCI individually yielded comparable results, but combining them improved prediction accuracy by approximately 10–15% across all drought categories, particularly with XGBoost. RF showed a similar trend, though its overall performance remained low. In contrast, combining ESI and DSCI did not enhance LSTM predictions; LSTM performed better when paired with DSCI alone.

Table 3: Accuracy of drought impact predictions at the state-level based on ESI, DSCI, and combined DSCI and ESI using XGBoost and testing data.

| New Mexico | Impacts (XGB's Results) -ESI- | | | | | | | | | |
|---|---|---|---|---|---|---|---|---|---|---|
| | Agriculture | | Water Supply & Quality | | Relief, Response & Restrictions | | Plants & Wildlife | | Fire | | Society & Public Health | |
| | F1-Score | | F1-Score | | F1-Score | | F1-Score | | F1-Score | | F1-Score | |
| Weeks - Period / Class | 0 | 1 | 0 | 1 | 0 | 1 | 0 | 1 | 0 | 1 | 0 | 1 |
| 1 : 8 | 0.89 | 0.86 | 0.85 | 0.79 | 0.90 | 0.87 | 0.81 | 0.69 | 0.94 | 0.94 | 0.81 | 0.70 |
| 2 : 8 | 0.84 | 0.76 | 0.87 | 0.82 | 0.91 | 0.88 | 0.81 | 0.70 | 0.95 | 0.94 | 0.91 | 0.89 |
| 3 : 8 | 0.81 | 0.70 | 0.90 | 0.88 | 0.91 | 0.89 | 0.82 | 0.73 | 0.92 | 0.90 | 0.88 | 0.85 |
| 4 : 8 | 0.89 | 0.86 | 0.85 | 0.78 | 0.90 | 0.88 | 0.81 | 0.71 | 0.87 | 0.83 | 0.79 | 0.63 |
| 5 : 8 | 0.86 | 0.80 | 0.81 | 0.68 | 0.87 | 0.82 | 0.79 | 0.64 | 0.88 | 0.85 | 0.73 | 0.43 |
| 6 : 8 | 0.87 | 0.82 | 0.83 | 0.74 | 0.87 | 0.83 | 0.78 | 0.62 | 0.81 | 0.69 | 0.79 | 0.63 |
| 7 : 8 | 0.82 | 0.73 | 0.79 | 0.65 | 0.84 | 0.78 | 0.77 | 0.60 | 0.81 | 0.71 | 0.97 | 0.97 |
| 8 : 8 | 0.74 | 0.51 | 0.76 | 0.55 | 0.76 | 0.60 | 0.76 | 0.56 | 0.75 | 0.55 | 0.80 | 0.78 |

| New Mexico | Impacts (XGB's Results) -DSCI- | | | | | | | | | |
|---|---|---|---|---|---|---|---|---|---|---|
| | Agriculture | | Water Supply & Quality | | Relief, Response & Restrictions | | Plants & Wildlife | | Fire | | Society & Public Health | |
| | F1-Score | | F1-Score | | F1-Score | | F1-Score | | F1-Score | | F1-Score | |
| Weeks - Period / Class | 0 | 1 | 0 | 1 | 0 | 1 | 0 | 1 | 0 | 1 | 0 | 1 |
| 1 : 8 | 0.88 | 0.84 | 0.93 | 0.92 | 0.93 | 0.91 | 0.84 | 0.77 | 0.95 | 0.94 | 0.89 | 0.87 |
| 2 : 8 | 0.86 | 0.81 | 0.91 | 0.89 | 0.90 | 0.88 | 0.83 | 0.74 | 0.92 | 0.91 | 0.86 | 0.80 |
| 3 : 8 | 0.87 | 0.82 | 0.85 | 0.78 | 0.90 | 0.87 | 0.79 | 0.61 | 0.94 | 0.93 | 0.75 | 0.52 |
| 4 : 8 | 0.84 | 0.77 | 0.88 | 0.84 | 0.92 | 0.90 | 0.84 | 0.76 | 0.90 | 0.87 | 0.78 | 0.62 |
| 5 : 8 | 0.85 | 0.79 | 0.88 | 0.85 | 0.89 | 0.85 | 0.75 | 0.50 | 0.90 | 0.88 | 0.85 | 0.79 |
| 6 : 8 | 0.84 | 0.77 | 0.83 | 0.74 | 0.85 | 0.80 | 0.89 | 0.86 | 0.85 | 0.79 | 0.81 | 0.68 |
| 7 : 8 | 0.83 | 0.75 | 0.80 | 0.68 | 0.86 | 0.80 | 0.84 | 0.76 | 0.81 | 0.71 | 0.79 | 0.65 |
| 8 : 8 | 0.77 | 0.60 | 0.76 | 0.51 | 0.79 | 0.66 | 0.79 | 0.66 | 0.81 | 0.70 | 0.83 | 0.75 |

| New Mexico | Impacts (XGB's Results) -DSCI & ESI- | | | | | | | | | |
|---|---|---|---|---|---|---|---|---|---|---|
| | Agriculture | | Water Supply & Quality | | Relief, Response & Restrictions | | Plants & Wildlife | | Fire | | Society & Public Health | |
| | F1-Score | | F1-Score | | F1-Score | | F1-Score | | F1-Score | | F1-Score | |
| Weeks - Period / Class | 0 | 1 | 0 | 1 | 0 | 1 | 0 | 1 | 0 | 1 | 0 | 1 |
| 1 : 8 | 0.90 | 0.87 | 0.88 | 0.83 | 0.95 | 0.95 | 0.87 | 0.82 | 0.94 | 0.94 | 0.82 | 0.71 |
| 2 : 8 | 0.91 | 0.88 | 0.89 | 0.85 | 0.96 | 0.95 | 0.86 | 0.81 | 0.95 | 0.95 | 0.89 | 0.89 |
| 3 : 8 | 0.93 | 0.92 | 0.89 | 0.85 | 0.95 | 0.95 | 0.89 | 0.86 | 0.92 | 0.90 | 0.89 | 0.86 |
| 4 : 8 | 0.94 | 0.93 | 0.81 | 0.70 | 0.92 | 0.91 | 0.89 | 0.87 | 0.90 | 0.87 | 0.81 | 0.68 |
| 5 : 8 | 0.84 | 0.77 | 0.87 | 0.82 | 0.93 | 0.92 | 0.85 | 0.79 | 0.90 | 0.88 | 0.92 | 0.91 |
| 6 : 8 | 0.90 | 0.87 | 0.82 | 0.73 | 0.94 | 0.93 | 0.94 | 0.93 | 0.89 | 0.86 | 0.80 | 0.67 |
| 7 : 8 | 0.86 | 0.81 | 0.79 | 0.65 | 0.92 | 0.90 | 0.86 | 0.80 | 0.89 | 0.87 | 0.91 | 0.89 |
| 8 : 8 | 0.83 | 0.74 | 0.78 | 0.63 | 0.91 | 0.89 | 0.82 | 0.72 | 0.83 | 0.74 | 0.77 | 0.59 |

Table 4: Accuracy of drought impact predictions at the state-level based on ESI, DSCI, and combined DSCI and ESI using LSTM and testing data.

| New Mexico | Impacts (LSTM's Results) -ESI- | | | | | | | | | | | |
|---|---|---|---|---|---|---|---|---|---|---|---|---|
| | Agriculture | | Water Supply & Quality | | Relief, Response & Restrictions | | Plants & Wildlife | | Fire | | Society & Public Health | |
| | F1-Score | | F1-Score | | F1-Score | | F1-Score | | F1-Score | | F1-Score | |
| Weeks - Period / Class | 0 | 1 | 0 | 1 | 0 | 1 | 0 | 1 | 0 | 1 | 0 | 1 |
| 1 : 8 | 0.70 | 0.47 | 0.75 | 0.62 | 0.79 | 0.76 | 0.75 | 0.57 | 0.87 | 0.86 | 0.70 | 0.32 |
| 2 : 8 | 0.74 | 0.64 | 0.68 | 0.39 | 0.75 | 0.68 | 0.70 | 0.40 | 0.77 | 0.64 | 0.66 | 0.12 |
| 3 : 8 | 0.70 | 0.46 | 0.69 | 0.44 | 0.73 | 0.63 | 0.68 | 0.30 | 0.70 | 0.45 | 0.69 | 0.25 |
| 4 : 8 | 0.66 | 0.28 | 0.72 | 0.52 | 0.73 | 0.60 | 0.64 | 0.12 | 0.74 | 0.61 | 0.65 | 0.02 |
| 5 : 8 | 0.71 | 0.52 | 0.69 | 0.39 | 0.73 | 0.66 | 0.65 | 0.16 | 0.76 | 0.64 | 0.73 | 0.47 |
| 6 : 8 | 0.70 | 0.56 | 0.72 | 0.54 | 0.69 | 0.59 | 0.63 | 0.03 | 0.69 | 0.56 | 0.65 | 0.10 |
| 7 : 8 | 0.68 | 0.49 | 0.71 | 0.49 | 0.74 | 0.74 | 0.70 | 0.51 | 0.67 | 0.52 | 0.65 | 0.07 |
| 8 : 8 | 0.65 | 0.55 | 0.67 | 0.45 | 0.66 | 0.64 | 0.66 | 0.15 | 0.65 | 0.63 | 0.70 | 0.37 |

| New Mexico | Impacts (LSTM's Results) -DSCI- | | | | | | | | | | | |
|---|---|---|---|---|---|---|---|---|---|---|---|---|
| | Agriculture | | Water Supply & Quality | | Relief, Response & Restrictions | | Plants & Wildlife | | Fire | | Society & Public Health | |
| | F1-Score | | F1-Score | | F1-Score | | F1-Score | | F1-Score | | F1-Score | |
| Weeks - Period / Class | 0 | 1 | 0 | 1 | 0 | 1 | 0 | 1 | 0 | 1 | 0 | 1 |
| 1 : 8 | 0.70 | 0.62 | 0.75 | 0.62 | 0.80 | 0.78 | 0.70 | 0.42 | 0.85 | 0.83 | 0.88 | 0.85 |
| 2 : 8 | 0.69 | 0.47 | 0.72 | 0.54 | 0.79 | 0.75 | 0.72 | 0.55 | 0.81 | 0.79 | 0.69 | 0.31 |
| 3 : 8 | 0.68 | 0.46 | 0.83 | 0.83 | 0.78 | 0.76 | 0.71 | 0.55 | 0.76 | 0.70 | 0.96 | 0.05 |
| 4 : 8 | 0.71 | 0.68 | 0.74 | 0.63 | 0.75 | 0.69 | 0.68 | 0.39 | 0.83 | 0.83 | 0.82 | 0.78 |
| 5 : 8 | 0.74 | 0.62 | 0.75 | 0.66 | 0.74 | 0.68 | 0.74 | 0.65 | 0.79 | 0.74 | 0.62 | 0.00 |
| 6 : 8 | 0.62 | 0.42 | 0.78 | 0.76 | 0.72 | 0.68 | 0.74 | 0.70 | 0.73 | 0.67 | 0.70 | 0.44 |
| 7 : 8 | 0.68 | 0.58 | 0.73 | 0.65 | 0.68 | 0.65 | 0.67 | 0.56 | 0.71 | 0.60 | 0.90 | 0.90 |
| 8 : 8 | 0.60 | 0.51 | 0.68 | 0.56 | 0.68 | 0.67 | 0.66 | 0.56 | 0.69 | 0.68 | 0.79 | 0.76 |

| New Mexico | Impacts (LSTM's Results) -DSCI & ESI- | | | | | | | | | | | |
|---|---|---|---|---|---|---|---|---|---|---|---|---|
| | Agriculture | | Water Supply & Quality | | Relief, Response & Restrictions | | Plants & Wildlife | | Fire | | Society & Public Health | |
| | F1-Score | | F1-Score | | F1-Score | | F1-Score | | F1-Score | | F1-Score | |
| Weeks - Period / Class | 0 | 1 | 0 | 1 | 0 | 1 | 0 | 1 | 0 | 1 | 0 | 1 |
| 1 : 8 | 0.74 | 0.64 | 0.72 | 0.50 | 0.77 | 0.70 | 0.78 | 0.70 | 0.81 | 0.75 | 0.85 | 0.80 |
| 2 : 8 | 0.66 | 0.45 | 0.73 | 0.55 | 0.75 | 0.69 | 0.77 | 0.70 | 0.76 | 0.66 | 0.87 | 0.85 |
| 3 : 8 | 0.69 | 0.56 | 0.70 | 0.51 | 0.74 | 0.66 | 0.82 | 0.77 | 0.78 | 0.73 | 0.73 | 0.52 |
| 4 : 8 | 0.65 | 0.33 | 0.73 | 0.58 | 0.76 | 0.70 | 0.68 | 0.40 | 0.75 | 0.62 | 0.69 | 0.32 |
| 5 : 8 | 0.67 | 0.46 | 0.69 | 0.41 | 0.75 | 0.70 | 0.74 | 0.65 | 0.72 | 0.56 | 0.87 | 0.85 |
| 6 : 8 | 0.66 | 0.40 | 0.68 | 0.43 | 0.76 | 0.72 | 0.73 | 0.60 | 0.74 | 0.65 | 0.84 | 0.79 |
| 7 : 8 | 0.71 | 0.56 | 0.71 | 0.53 | 0.73 | 0.65 | 0.74 | 0.63 | 0.73 | 0.66 | 0.73 | 0.51 |
| 8 : 8 | 0.62 | 0.33 | 0.69 | 0.44 | 0.75 | 0.71 | 0.66 | 0.42 | 0.68 | 0.56 | 0.66 | 0.21 |

Table 5: Accuracy of drought impact predictions at the state-level based on ESI, DSCI, and combined DSCI and ESI using RF and testing data.

| New Mexico | Impacts (RF's Results) -ESI- | | | | | | | | | | | |
|---|---|---|---|---|---|---|---|---|---|---|---|---|
| | Agriculture | | Water Supply & Quality | | Relief, Response & Restrictions | | Plants & Wildlife | | Fire | | Society & Public Health | |
| | F1-Score | | F1-Score | | F1-Score | | F1-Score | | F1-Score | | F1-Score | |
| Weeks - Period / Class | 0 | 1 | 0 | 1 | 0 | 1 | 0 | 1 | 0 | 1 | 0 | 1 |
| 1 : 8 | 0.74 | 0.44 | 0.72 | 0.38 | 0.84 | 0.76 | 0.70 | 0.22 | 0.83 | 0.74 | 0.71 | 0.33 |
| 2 : 8 | 0.73 | 0.43 | 0.76 | 0.53 | 0.80 | 0.65 | 0.67 | 0.05 | 0.80 | 0.65 | 0.71 | 0.33 |
| 3 : 8 | 0.76 | 0.54 | 0.70 | 0.26 | 0.80 | 0.66 | 0.68 | 0.13 | 0.81 | 0.68 | 0.68 | 0.07 |
| 4 : 8 | 0.75 | 0.50 | 0.73 | 0.41 | 0.78 | 0.59 | 0.77 | 0.58 | 0.79 | 0.63 | 0.67 | 0.00 |
| 5 : 8 | 0.72 | 0.35 | 0.72 | 0.37 | 0.78 | 0.59 | 0.69 | 0.18 | 0.78 | 0.60 | 0.70 | 0.00 |
| 6 : 8 | 0.75 | 0.49 | 0.72 | 0.34 | 0.75 | 0.49 | 0.71 | 0.29 | 0.76 | 0.55 | 0.67 | 0.00 |
| 7 : 8 | 0.71 | 0.28 | 0.68 | 0.09 | 0.73 | 0.42 | 0.69 | 0.15 | 0.76 | 0.52 | 0.67 | 0.00 |
| 8 : 8 | 0.73 | 0.43 | 0.74 | 0.46 | 0.73 | 0.40 | 0.67 | 0.04 | 0.68 | 0.09 | 0.67 | 0.00 |

| New Mexico | Impacts (RF's Results) -DSCI- | | | | | | | | | | | |
|---|---|---|---|---|---|---|---|---|---|---|---|---|
| | Agriculture | | Water Supply & Quality | | Relief, Response & Restrictions | | Plants & Wildlife | | Fire | | Society & Public Health | |
| | F1-Score | | F1-Score | | F1-Score | | F1-Score | | F1-Score | | F1-Score | |
| Weeks - Period / Class | 0 | 1 | 0 | 1 | 0 | 1 | 0 | 1 | 0 | 1 | 0 | 1 |
| 1 : 8 | 0.76 | 0.54 | 0.76 | 0.54 | 0.85 | 0.79 | 0.70 | 0.25 | 0.85 | 0.78 | 0.68 | 0.10 |
| 2 : 8 | 0.76 | 0.53 | 0.72 | 0.38 | 0.86 | 0.80 | 0.70 | 0.27 | 0.87 | 0.82 | 0.68 | 0.11 |
| 3 : 8 | 0.73 | 0.39 | 0.71 | 0.29 | 0.85 | 0.79 | 0.77 | 0.58 | 0.83 | 0.74 | 0.72 | 0.34 |
| 4 : 8 | 0.72 | 0.34 | 0.72 | 0.33 | 0.80 | 0.66 | 0.75 | 0.50 | 0.80 | 0.66 | 0.67 | 0.05 |
| 5 : 8 | 0.73 | 0.39 | 0.74 | 0.43 | 0.78 | 0.60 | 0.71 | 0.32 | 0.76 | 0.52 | 0.70 | 0.26 |
| 6 : 8 | 0.74 | 0.46 | 0.75 | 0.49 | 0.77 | 0.56 | 0.72 | 0.32 | 0.78 | 0.59 | 0.67 | 0.00 |
| 7 : 8 | 0.72 | 0.37 | 0.77 | 0.58 | 0.79 | 0.63 | 0.70 | 0.24 | 0.77 | 0.58 | 0.70 | 0.22 |
| 8 : 8 | 0.74 | 0.45 | 0.76 | 0.55 | 0.83 | 0.73 | 0.73 | 0.34 | 0.77 | 0.56 | 0.70 | 0.25 |

| New Mexico | Impacts (RF's Results) -DSCI & ESI- | | | | | | | | | | | |
|---|---|---|---|---|---|---|---|---|---|---|---|---|
| | Agriculture | | Water Supply & Quality | | Relief, Response & Restrictions | | Plants & Wildlife | | Fire | | Society & Public Health | |
| | F1-Score | | F1-Score | | F1-Score | | F1-Score | | F1-Score | | F1-Score | |
| Weeks - Period / Class | 0 | 1 | 0 | 1 | 0 | 1 | 0 | 1 | 0 | 1 | 0 | 1 |
| 1 : 8 | 0.75 | 0.49 | 0.74 | 0.45 | 0.86 | 0.80 | 0.78 | 0.62 | 0.87 | 0.83 | 0.72 | 0.38 |
| 2 : 8 | 0.75 | 0.49 | 0.76 | 0.53 | 0.84 | 0.77 | 0.72 | 0.35 | 0.82 | 0.71 | 0.73 | 0.42 |
| 3 : 8 | 0.75 | 0.50 | 0.68 | 0.10 | 0.86 | 0.81 | 0.75 | 0.50 | 0.82 | 0.71 | 0.73 | 0.40 |
| 4 : 8 | 0.78 | 0.60 | 0.74 | 0.47 | 0.84 | 0.77 | 0.71 | 0.33 | 0.82 | 0.72 | 0.70 | 0.23 |
| 5 : 8 | 0.78 | 0.62 | 0.75 | 0.50 | 0.79 | 0.63 | 0.76 | 0.55 | 0.82 | 0.71 | 0.71 | 0.29 |
| 6 : 8 | 0.74 | 0.46 | 0.74 | 0.47 | 0.82 | 0.71 | 0.73 | 0.39 | 0.80 | 0.67 | 0.74 | 0.46 |
| 7 : 8 | 0.77 | 0.58 | 0.74 | 0.46 | 0.80 | 0.65 | 0.75 | 0.50 | 0.73 | 0.42 | 0.67 | 0.00 |
| 8 : 8 | 0.78 | 0.62 | 0.69 | 0.17 | 0.79 | 0.62 | 0.74 | 0.46 | 0.77 | 0.57 | 0.67 | 0.02 |

### 4.3. County-Level Predictions

Model performance at the county level is illustrated for Bernalillo County (Tables 6 – 8). At the state level, none of the models predicted Business or Energy impacts due to a lack of convergence; these categories were excluded. XGBoost showed strong accuracy across most impact categories and prior-week data, except for Water and Society when using ESI and DSCI. Most F1-scores ranged from 0.70 – 0.98, with a few between 0.50 – 0.69; scores below 0.50 were considered unacceptable. LSTM and RF performed poorly, with most scores below 0.50 and only occasional values above 0.50 when combining ESI and DSCI. Among categories, Agriculture and Fire had the highest scores (0.96 – 0.73), followed by Relief (0.93 – 0.78), then Water and Plants (0.77 – 0.59). Extending prior weeks reduced accuracy, but even using only the eighth prior week, XGBoost maintained F1-scores above 0.50 for Agriculture, Fire, Plants, Relief, Response, Restrictions, and Society, and slightly below 0.50 for Water.

Table 6: Accuracy of drought impact predictions at the county-level based on ESI, DSCI, and combined DSCI and ESI using XGBoost and testing data over Bernalillo County.

| Bernalillo | Impacts (XGB's Results) -ESI- | | | | | | | | | | | |
|---|---|---|---|---|---|---|---|---|---|---|---|---|
| | Agriculture | | Water Supply & Quality | | Relief, Response & Restrictions | | Plants & Wildlife | | Fire | | Society & Public Health | |
| | F1-Score | | F1-Score | | F1-Score | | F1-Score | | F1-Score | | F1-Score | |
| Weeks - Period / Class | 0 | 1 | 0 | 1 | 0 | 1 | 0 | 1 | 0 | 1 | 0 | 1 |
| 1:8 | 0.95 | 0.94 | 0.88 | 0.85 | 0.93 | 0.91 | 0.85 | 0.78 | 0.95 | 0.95 | 0.80 | 0.67 |
| 2:8 | 0.94 | 0.94 | 0.84 | 0.77 | 0.87 | 0.83 | 0.88 | 0.84 | 0.94 | 0.93 | 0.91 | 0.90 |
| 3:8 | 0.87 | 0.83 | 0.88 | 0.83 | 0.86 | 0.80 | 0.88 | 0.84 | 0.92 | 0.91 | 0.86 | 0.81 |
| 4:8 | 0.87 | 0.83 | 0.82 | 0.71 | 0.84 | 0.77 | 0.87 | 0.83 | 0.92 | 0.91 | 0.75 | 0.50 |
| 5:8 | 0.86 | 0.81 | 0.83 | 0.75 | 0.86 | 0.80 | 0.79 | 0.65 | 0.86 | 0.80 | 0.71 | 0.32 |
| 6:8 | 0.91 | 0.89 | 0.79 | 0.64 | 0.81 | 0.70 | 0.84 | 0.78 | 0.81 | 0.71 | 0.78 | 0.61 |
| 7:8 | 0.93 | 0.90 | 0.85 | 0.77 | 0.81 | 0.69 | 0.79 | 0.65 | 0.86 | 0.80 | 0.87 | 0.82 |
| 8:8 | 0.76 | 0.54 | 0.77 | 0.52 | 0.81 | 0.67 | 0.80 | 0.66 | 0.79 | 0.65 | 0.81 | 0.69 |

| Bernalillo | Impacts (XGB's Results) -DSCI- | | | | | | | | | | | |
|---|---|---|---|---|---|---|---|---|---|---|---|---|
| | Agriculture | | Water Supply & Quality | | Relief, Response & Restrictions | | Plants & Wildlife | | Fire | | Society & Public Health | |
| | F1-Score | | F1-Score | | F1-Score | | F1-Score | | F1-Score | | F1-Score | |
| Weeks - Period / Class | 0 | 1 | 0 | 1 | 0 | 1 | 0 | 1 | 0 | 1 | 0 | 1 |
| 1:8 | 0.93 | 0.92 | 0.83 | 0.75 | 0.91 | 0.88 | 0.80 | 0.67 | 0.95 | 0.95 | 0.82 | 0.71 |
| 2:8 | 0.95 | 0.95 | 0.84 | 0.76 | 0.88 | 0.84 | 0.94 | 0.77 | 0.87 | 0.83 | 0.88 | 0.84 |
| 3:8 | 0.89 | 0.86 | 0.83 | 0.73 | 0.82 | 0.73 | 0.86 | 0.80 | 0.84 | 0.77 | 0.84 | 0.77 |
| 4:8 | 0.85 | 0.79 | 0.82 | 0.71 | 0.85 | 0.78 | 0.80 | 0.66 | 0.85 | 0.79 | 0.76 | 0.55 |
| 5:8 | 0.82 | 0.71 | 0.80 | 0.66 | 0.82 | 0.72 | 0.80 | 0.65 | 0.80 | 0.67 | 0.78 | 0.59 |
| 6:8 | 0.75 | 0.52 | 0.82 | 0.71 | 0.81 | 0.68 | 0.81 | 0.68 | 0.80 | 0.64 | 0.75 | 0.50 |
| 7:8 | 0.79 | 0.62 | 0.78 | 0.56 | 0.80 | 0.64 | 0.80 | 0.66 | 0.79 | 0.63 | 0.70 | 0.24 |
| 8:8 | 0.80 | 0.66 | 0.77 | 0.47 | 0.81 | 0.65 | 0.80 | 0.67 | 0.76 | 0.52 | 0.78 | 0.59 |

| Bernalillo | Impacts (XGB's Results) -DSCI & ESI- | | | | | | | | | | | |
|---|---|---|---|---|---|---|---|---|---|---|---|---|
| | Agriculture | | Water Supply & Quality | | Relief, Response & Restrictions | | Plants & Wildlife | | Fire | | Society & Public Health | |
| | F1-Score | | F1-Score | | F1-Score | | F1-Score | | F1-Score | | F1-Score | |
| Weeks - Period / Class | 0 | 1 | 0 | 1 | 0 | 1 | 0 | 1 | 0 | 1 | 0 | 1 |
| 1:8 | 0.97 | 0.96 | 0.84 | 0.77 | 0.94 | 0.93 | 0.83 | 0.75 | 0.95 | 0.95 | 0.85 | 0.79 |
| 2:8 | 0.95 | 0.94 | 0.89 | 0.85 | 0.92 | 0.90 | 0.86 | 0.81 | 0.94 | 0.94 | 0.80 | 0.67 |
| 3:8 | 0.99 | 0.99 | 0.90 | 0.87 | 0.93 | 0.92 | 0.88 | 0.84 | 0.93 | 0.92 | 0.83 | 0.75 |
| 4:8 | 0.97 | 0.97 | 0.86 | 0.80 | 0.92 | 0.91 | 0.85 | 0.78 | 0.93 | 0.92 | 0.79 | 0.64 |
| 5:8 | 0.98 | 0.98 | 0.83 | 0.74 | 0.94 | 0.93 | 0.86 | 0.81 | 0.91 | 0.88 | 0.77 | 0.57 |
| 6:8 | 0.95 | 0.95 | 0.89 | 0.85 | 0.93 | 0.91 | 0.86 | 0.79 | 0.91 | 0.89 | 0.77 | 0.57 |
| 7:8 | 0.85 | 0.79 | 0.87 | 0.82 | 0.93 | 0.91 | 0.85 | 0.79 | 0.90 | 0.87 | 0.81 | 0.68 |
| 8:8 | 0.85 | 0.78 | 0.82 | 0.69 | 0.86 | 0.78 | 0.78 | 0.59 | 0.83 | 0.73 | 0.84 | 0.76 |

Table 7: Accuracy of drought impact predictions at the county-level based on ESI, DSCI, and combined DSCI and ESI using LSTM and testing data over Bernalillo County.

**Impacts (LSTM's Results) -ESI-**

| Bernalillo | Agriculture | | Water Supply & Quality | | Relief, Response & Restrictions | | Plants & Wildlife | | Fire | | Society & Public Health | |
|---|---|---|---|---|---|---|---|---|---|---|---|---|
| | F1-Score | | F1-Score | | F1-Score | | F1-Score | | F1-Score | | F1-Score | |
| Weeks - Period / Class | 0 | 1 | 0 | 1 | 0 | 1 | 0 | 1 | 0 | 1 | 0 | 1 |
| 1 : 8 | 0.69 | 0.33 | 0.70 | 0.24 | 0.74 | 0.49 | 0.71 | 0.39 | 0.74 | 0.49 | 0.67 | 0.00 |
| 2 : 8 | 0.74 | 0.53 | 0.69 | 0.20 | 0.72 | 0.42 | 0.75 | 0.56 | 0.74 | 0.47 | 0.66 | 0.00 |
| 3 : 8 | 0.68 | 0.29 | 0.67 | 0.07 | 0.69 | 0.25 | 0.67 | 0.21 | 0.67 | 0.11 | 0.69 | 0.23 |
| 4 : 8 | 0.72 | 0.49 | 0.68 | 0.08 | 0.68 | 0.28 | 0.65 | 0.12 | 0.66 | 0.08 | 0.67 | 0.03 |
| 5 : 8 | 0.69 | 0.37 | 0.66 | 0.04 | 0.67 | 0.14 | 0.65 | 0.11 | 0.67 | 0.14 | 0.66 | 0.00 |
| 6 : 8 | 0.67 | 0.29 | 0.66 | 0.07 | 0.66 | 0.07 | 0.65 | 0.32 | 0.65 | 0.02 | 0.67 | 0.24 |
| 7 : 8 | 0.77 | 0.70 | 0.67 | 0.06 | 0.67 | 0.20 | 0.65 | 0.13 | 0.67 | 0.17 | 0.66 | 0.00 |
| 8 : 8 | 0.69 | 0.31 | 0.69 | 0.21 | 0.66 | 0.20 | 0.70 | 0.34 | 0.65 | 0.06 | 0.67 | 0.00 |

**Impacts (LSTM's Results) -DSCI-**

| Bernalillo | Agriculture | | Water Supply & Quality | | Relief, Response & Restrictions | | Plants & Wildlife | | Fire | | Society & Public Health | |
|---|---|---|---|---|---|---|---|---|---|---|---|---|
| | F1-Score | | F1-Score | | F1-Score | | F1-Score | | F1-Score | | F1-Score | |
| Weeks - Period / Class | 0 | 1 | 0 | 1 | 0 | 1 | 0 | 1 | 0 | 1 | 0 | 1 |
| 1 : 8 | 0.70 | 0.38 | 0.68 | 0.15 | 0.71 | 0.45 | 0.78 | 0.64 | 0.77 | 0.58 | 0.74 | 0.48 |
| 2 : 8 | 0.73 | 0.49 | 0.69 | 0.21 | 0.71 | 0.33 | 0.76 | 0.60 | 0.72 | 0.38 | 0.66 | 0.03 |
| 3 : 8 | 0.74 | 0.54 | 0.69 | 0.21 | 0.71 | 0.25 | 0.71 | 0.42 | 0.70 | 0.36 | 0.60 | 0.00 |
| 4 : 8 | 0.67 | 0.28 | 0.68 | 0.13 | 0.70 | 0.22 | 0.66 | 0.35 | 0.70 | 0.28 | 0.66 | 0.01 |
| 5 : 8 | 0.69 | 0.30 | 0.69 | 0.08 | 0.70 | 0.16 | 0.67 | 0.00 | 0.67 | 0.01 | 0.66 | 0.00 |
| 6 : 8 | 0.67 | 0.17 | 0.69 | 0.09 | 0.69 | 0.25 | 0.65 | 0.00 | 0.66 | 0.15 | 0.65 | 0.08 |
| 7 : 8 | 0.67 | 0.21 | 0.68 | 0.00 | 0.70 | 0.30 | 0.65 | 0.07 | 0.66 | 0.00 | 0.64 | 0.01 |
| 8 : 8 | 0.72 | 0.58 | 0.68 | 0.09 | 0.70 | 0.09 | 0.62 | 0.05 | 0.63 | 0.17 | 0.65 | 0.07 |

**Impacts (LSTM's Results) -DSCI & ESI-**

| Bernalillo | Agriculture | | Water Supply & Quality | | Relief, Response & Restrictions | | Plants & Wildlife | | Fire | | Society & Public Health | |
|---|---|---|---|---|---|---|---|---|---|---|---|---|
| | F1-Score | | F1-Score | | F1-Score | | F1-Score | | F1-Score | | F1-Score | |
| Weeks - Period / Class | 0 | 1 | 0 | 1 | 0 | 1 | 0 | 1 | 0 | 1 | 0 | 1 |
| 1 : 8 | 0.68 | 0.40 | 0.73 | 0.50 | 0.75 | 0.58 | 0.78 | 0.67 | 0.73 | 0.51 | 0.66 | 0.01 |
| 2 : 8 | 0.72 | 0.54 | 0.69 | 0.23 | 0.72 | 0.47 | 0.75 | 0.59 | 0.75 | 0.54 | 0.66 | 0.00 |
| 3 : 8 | 0.67 | 0.33 | 0.66 | 0.14 | 0.70 | 0.43 | 0.71 | 0.46 | 0.72 | 0.38 | 0.66 | 0.00 |
| 4 : 8 | 0.70 | 0.40 | 0.67 | 0.24 | 0.68 | 0.28 | 0.69 | 0.31 | 0.71 | 0.38 | 0.66 | 0.02 |
| 5 : 8 | 0.68 | 0.41 | 0.65 | 0.03 | 0.67 | 0.25 | 0.66 | 0.18 | 0.68 | 0.17 | 0.66 | 0.00 |
| 6 : 8 | 0.71 | 0.53 | 0.66 | 0.09 | 0.68 | 0.33 | 0.69 | 0.32 | 0.67 | 0.22 | 0.66 | 0.05 |
| 7 : 8 | 0.69 | 0.39 | 0.65 | 0.02 | 0.70 | 0.37 | 0.67 | 0.22 | 0.63 | 0.00 | 0.66 | 0.02 |
| 8 : 8 | 0.69 | 0.52 | 0.71 | 0.38 | 0.70 | 0.42 | 0.66 | 0.27 | 0.65 | 0.04 | 0.68 | 0.21 |

Table 8: Accuracy of drought impact predictions at the county-level based on ESI, DSCI, and combined DSCI and ESI using RF and testing data over Bernalillo County.

**Impacts (RF's Results) -ESI-**

| Bernalillo | Agriculture | | Water Supply & Quality | | Relief, Response & Restrictions | | Plants & Wildlife | | Fire | | Society & Public Health | |
|---|---|---|---|---|---|---|---|---|---|---|---|---|
| | F1-Score | | F1-Score | | F1-Score | | F1-Score | | F1-Score | | F1-Score | |
| Weeks - Period / Class | 0 | 1 | 0 | 1 | 0 | 1 | 0 | 1 | 0 | 1 | 0 | 1 |
| 1 : 8 | 0.71 | 0.31 | 0.67 | 0.00 | 0.74 | 0.41 | 0.67 | 0.00 | 0.70 | 0.23 | 0.67 | 0.00 |
| 2 : 8 | 0.73 | 0.42 | 0.67 | 0.00 | 0.74 | 0.46 | 0.67 | 0.00 | 0.73 | 0.43 | 0.67 | 0.00 |
| 3 : 8 | 0.71 | 0.33 | 0.67 | 0.00 | 0.70 | 0.22 | 0.67 | 0.00 | 0.72 | 0.35 | 0.67 | 0.00 |
| 4 : 8 | 0.72 | 0.33 | 0.67 | 0.00 | 0.70 | 0.20 | 0.67 | 0.00 | 0.70 | 0.21 | 0.67 | 0.00 |
| 5 : 8 | 0.70 | 0.24 | 0.67 | 0.00 | 0.69 | 0.13 | 0.67 | 0.00 | 0.67 | 0.00 | 0.67 | 0.00 |
| 6 : 8 | 0.71 | 0.31 | 0.67 | 0.00 | 0.70 | 0.19 | 0.67 | 0.00 | 0.69 | 0.15 | 0.68 | 0.05 |
| 7 : 8 | 0.73 | 0.41 | 0.67 | 0.00 | 0.69 | 0.09 | 0.68 | 0.00 | 0.67 | 0.03 | 0.67 | 0.00 |
| 8 : 8 | 0.70 | 0.22 | 0.68 | 0.01 | 0.70 | 0.07 | 0.68 | 0.00 | 0.68 | 0.05 | 0.67 | 0.00 |

**Impacts (RF's Results) -DSCI-**

| Bernalillo | Agriculture | | Water Supply & Quality | | Relief, Response & Restrictions | | Plants & Wildlife | | Fire | | Society & Public Health | |
|---|---|---|---|---|---|---|---|---|---|---|---|---|
| | F1-Score | | F1-Score | | F1-Score | | F1-Score | | F1-Score | | F1-Score | |
| Weeks - Period / Class | 0 | 1 | 0 | 1 | 0 | 1 | 0 | 1 | 0 | 1 | 0 | 1 |
| 1 : 8 | 0.71 | 0.26 | 0.67 | 0.00 | 0.73 | 0.37 | 0.67 | 0.00 | 0.71 | 0.32 | 0.67 | 0.00 |
| 2 : 8 | 0.72 | 0.33 | 0.67 | 0.00 | 0.73 | 0.41 | 0.67 | 0.01 | 0.73 | 0.40 | 0.67 | 0.00 |
| 3 : 8 | 0.71 | 0.32 | 0.67 | 0.00 | 0.70 | 0.22 | 0.67 | 0.00 | 0.71 | 0.30 | 0.67 | 0.00 |
| 4 : 8 | 0.69 | 0.18 | 0.67 | 0.00 | 0.70 | 0.21 | 0.67 | 0.00 | 0.70 | 0.21 | 0.67 | 0.00 |
| 5 : 8 | 0.70 | 0.25 | 0.68 | 0.00 | 0.69 | 0.12 | 0.67 | 0.00 | 0.67 | 0.00 | 0.67 | 0.00 |
| 6 : 8 | 0.70 | 0.22 | 0.68 | 0.00 | 0.70 | 0.20 | 0.67 | 0.00 | 0.68 | 0.01 | 0.67 | 0.00 |
| 7 : 8 | 0.70 | 0.18 | 0.68 | 0.00 | 0.70 | 0.16 | 0.68 | 0.01 | 0.68 | 0.05 | 0.67 | 0.00 |
| 8 : 8 | 0.71 | 0.25 | 0.69 | 0.00 | 0.70 | 0.08 | 0.67 | 0.01 | 0.68 | 0.00 | 0.68 | 0.00 |

**Impacts (RF's Results) -DSCI & ESI-**

| Bernalillo | Agriculture | | Water Supply & Quality | | Relief, Response & Restrictions | | Plants & Wildlife | | Fire | | Society & Public Health | |
|---|---|---|---|---|---|---|---|---|---|---|---|---|
| | F1-Score | | F1-Score | | F1-Score | | F1-Score | | F1-Score | | F1-Score | |
| Weeks - Period / Class | 0 | 1 | 0 | 1 | 0 | 1 | 0 | 1 | 0 | 1 | 0 | 1 |
| 1 : 8 | 0.77 | 0.57 | 0.67 | 0.01 | 0.84 | 0.76 | 0.68 | 0.09 | 0.77 | 0.55 | 0.67 | 0.01 |
| 2 : 8 | 0.76 | 0.53 | 0.67 | 0.00 | 0.81 | 0.67 | 0.67 | 0.01 | 0.78 | 0.58 | 0.67 | 0.00 |
| 3 : 8 | 0.81 | 0.70 | 0.67 | 0.01 | 0.81 | 0.67 | 0.67 | 0.03 | 0.76 | 0.46 | 0.67 | 0.00 |
| 4 : 8 | 0.75 | 0.48 | 0.67 | 0.00 | 0.77 | 0.54 | 0.67 | 0.02 | 0.74 | 0.43 | 0.67 | 0.00 |
| 5 : 8 | 0.76 | 0.52 | 0.67 | 0.00 | 0.76 | 0.49 | 0.67 | 0.00 | 0.69 | 0.18 | 0.68 | 0.06 |
| 6 : 8 | 0.73 | 0.40 | 0.67 | 0.00 | 0.75 | 0.48 | 0.67 | 0.03 | 0.70 | 0.22 | 0.67 | 0.00 |
| 7 : 8 | 0.75 | 0.49 | 0.68 | 0.00 | 0.73 | 0.34 | 0.67 | 0.00 | 0.69 | 0.11 | 0.67 | 0.05 |
| 8 : 8 | 0.76 | 0.51 | 0.68 | 0.02 | 0.73 | 0.33 | 0.68 | 0.02 | 0.68 | 0.00 | 0.69 | 0.13 |

## 5. Discussion

This analysis demonstrated that the XGBoost model, when combined with data augmentation and spatial-temporal features, can successfully predict specific drought impacts up to eight weeks in advance. The superior performance of XGBoost over RF and LSTM is likely due to its native ability to handle the sparse and imbalanced dataset, particularly the missing values introduced by the neighboring-county feature engineering.

### 5.1. Predictability of Impacts and Index Performance

The evaluation of RF, LSTM, and XGBoost revealed distinct limitations in handling sparse and structurally inconsistent data. RF, as a classical ML approach, struggled with limited feature representation—reported impacts accounted for only ~5% of all features—and with missing data caused by varying numbers of neighboring counties. LSTM performed better but was constrained by short time series, suggesting deeper architectures could improve accuracy. XGBoost consistently outperformed both models, effectively managing limited instances, structural variability, and sparse impact features.

The model's performance varied significantly across impact categories, revealing insights into the nature of the impacts themselves. At both the state (Figure 6) and county (Figure 7) scales, **Fire** and **Relief** impacts exhibited the highest consistency and the most significant predictive accuracy. This result is reasonable, as these impact types correspond to discrete, well-documented events (*e.g.*, wildfires or disaster declarations), providing the model with a strong and unambiguous signal. **Agriculture** and **Water** impact also exhibited strong predictability, likely due to their direct association with the physical processes represented in the drought indices.

Conversely, **Society & Public Health** impacts were predicted with low and inconsistent accuracy. This aligns with some previous findings of Noel *et al.* (2020) and can be explained by two factors: 1) Societal impacts are often a *cascading* effect of other impacts (*e.g.*, agricultural losses leading to economic stress) and are not a direct physical response; and 2) These impacts are highly subject to temporal and spatial reporting biases.

The choice of drought index was also critical. The combined **DSCI & ESI** index provided the most robust predictions for physical-ecosystem impacts (Agriculture, Plants, Fire). This combination is effective because the DSCI incorporates quasi-observational, on-the-ground expert assessments. At the same time, the ESI captures the rapid-onset physical stress associated with "flash droughts" (*e.g.*, vegetation stress, soil moisture deficit). Interestingly, the **DSCI alone** often provided the lowest performance, especially at the county level (Figure 7). This may be because the DSCI's reliance on human expert input, while valuable at a macro level, can be "after-the-fact" and may not capture the fine-scale, immediate physical signals that the ESI detects.

County-level results mirrored state-level patterns (Figure 7). **DSCI & ESI** provided the highest accuracy for Agriculture, Fire, and Relief, followed by ESI and then DSCI. ESI better captured plant impacts due to its sensitivity to soil moisture and vegetation stress. Society impacts remained difficult to predict because of spatial and temporal reporting inconsistencies. Overall, DSCI alone yielded the weakest performance, partly because it relies heavily on human observations, which often reflect conditions after impacts occur. Future improvements should focus on integrating multiple indices and enhancing temporal and spatial precision in impact reporting.

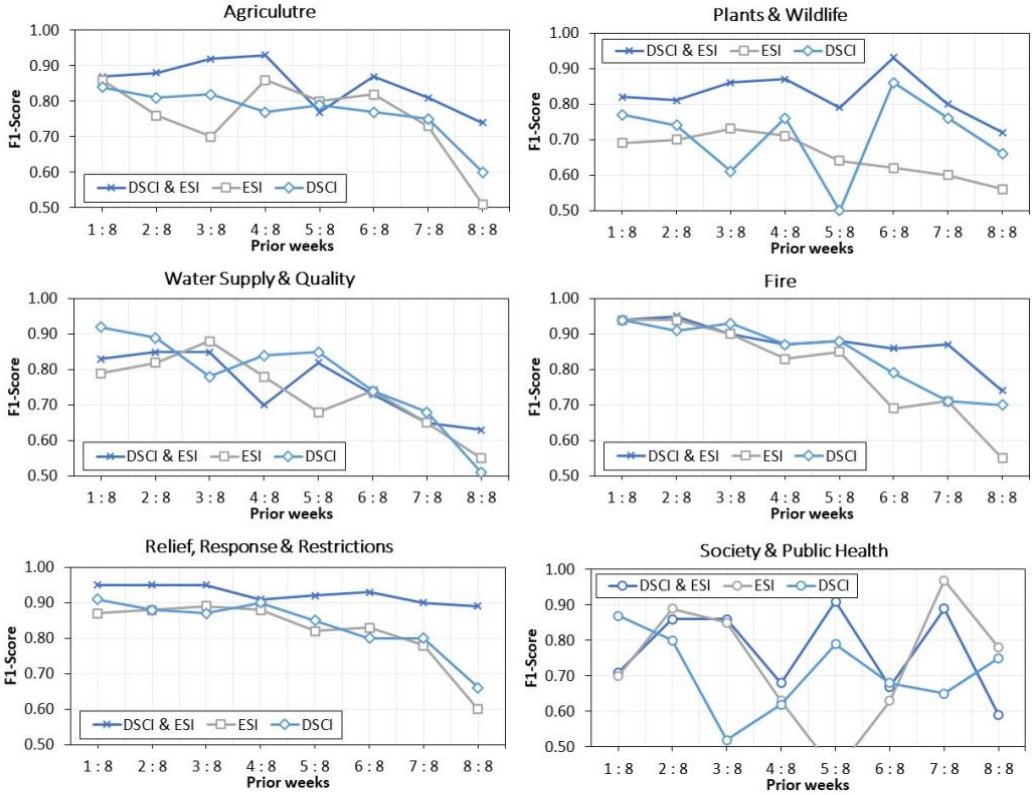

Figure 6: Summary of drought impact predictions based on the XGBoost model at the state-level. Each panel shows the performance of the different drought indices, including the DSCI, ESI, and their combined use.

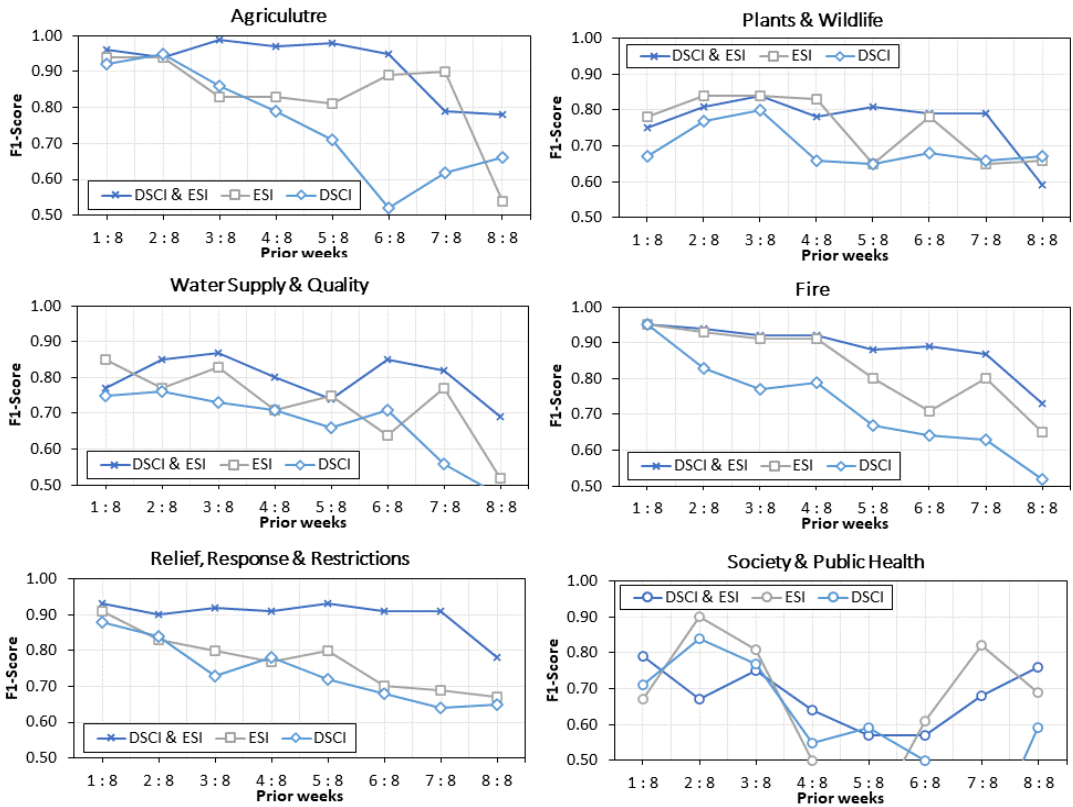

Figure 7: Summary of drought impact predictions based on the XGBoost model at the county level. Each panel shows the performance of the different drought indices, including the DSCI, ESI, and their combined use over Bernalillo County.

## 5.2. Interpretation of Feature Contribution

The feature contribution evaluation was used to compare the models' performance, understand how the different features (*i.e.*, variables) are represented, understand whether this information can be used to highlight any issues with the data, and provide guidance in drought impacts data collection.

The XGBoost models demonstrated a consistent pattern in feature representation (Figures 8 & 9). Drought impacts from neighboring counties contributed the most at 66%, followed by the neighboring counties' drought index at 21%. In comparison, the target county's impacts accounted for 11% and only 2% for its drought index. RF models showed a similar tendency, but with slightly lower contributions from neighboring counties, at 57% for impacts and 17% for drought indices. In comparison, the target county's impacts accounted for 21% and 5% for its index. Overall, neighboring counties accounted for 87% in XGBoost's and 74% in RF, leaving the target county at 13% and 26%, respectively. Unlike XGBoost, RF's feature contributions were not consistent throughout all prior weeks and varied significantly.

This observation can be explained in two ways:
1. **Physical Reality:** Drought is a regional, spatially continuous phenomenon that does not adhere to administrative boundaries. Therefore, the drought conditions and impacts in an adjacent county are, in fact, highly correlated and strongly predictive of conditions in the target county.
2. **Data Volume:** This result is also influenced by feature-engineering artifacts. Each target county has a variable number of neighbors (at least 4). By including 8 weeks of data for multiple indices and impacts from all neighbors, the *volume* of feature data from neighbors is overwhelmingly larger than the volume of data from the single target county. The model, therefore, found more information and predictive power in this much larger pool of spatially correlated data.

This finding suggests that while the model is effective, its predictive power relies heavily on this spatial augmentation. Future work could aim to enrich the *target county's* data by including more diverse indices to see if its local feature contribution can be increased.

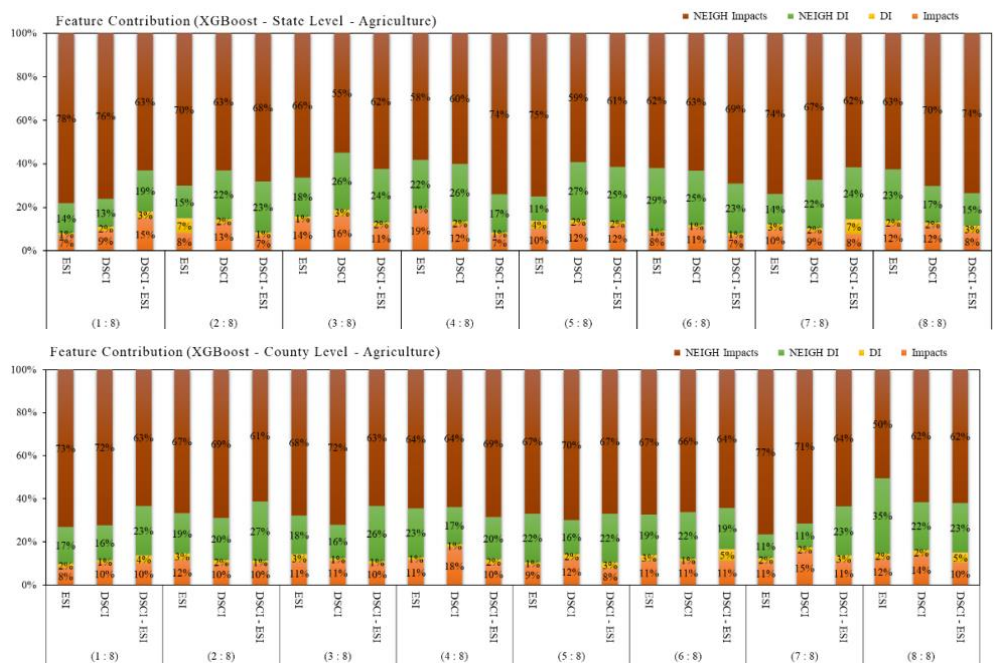

Figure 8: Weight of contributions based on the XGBoost at the state and county levels for the four different groups of features used in the modeling, including the drought impacts, drought index, neighboring counties impacts, and neighboring counties drought index.

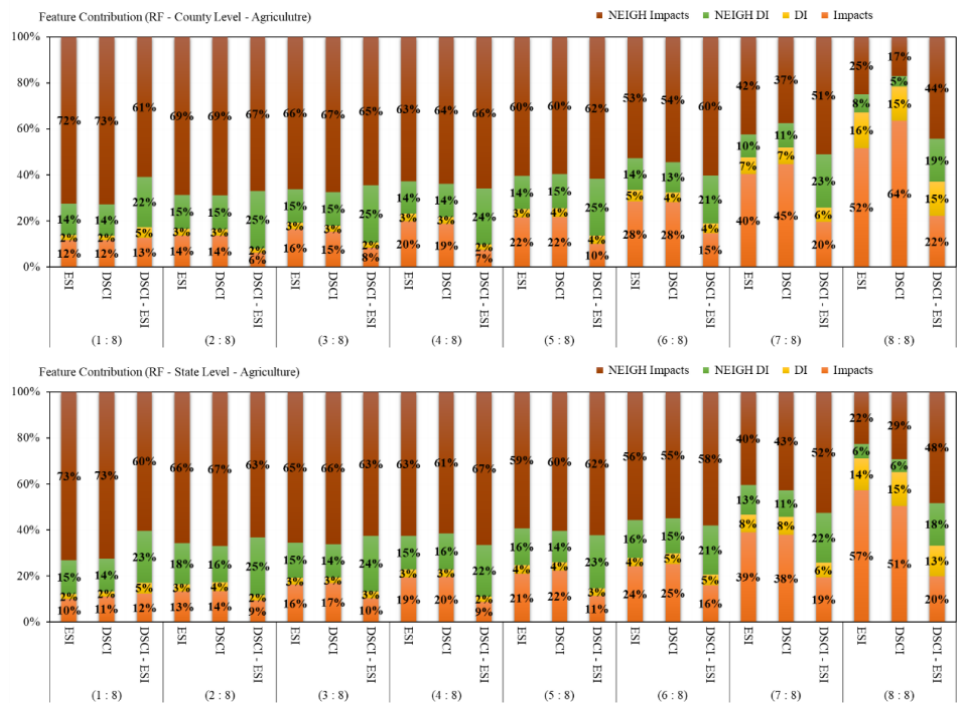

Figure 9: Weight of contributions based on the RF at the state and county levels for the four different groups of features used in the modeling, including the drought impacts, drought index, neighboring counties impacts, and neighboring counties drought index.

### 5.3. Forecasting Implications

To support farmers, resource managers, and policymakers, providing drought impact forecasts with lead times is critical for adaptation and mitigation. The retrospective approach used here can be applied operationally. For example, forecasts for June 3, 2024, were generated using data from April 8 – May 27, offering a one-week lead time. Shortening the historical window by one week increases lead time to two weeks, and using only April 8 data can provide an eight-week lead time. Forecasts for June 3 ranged from 2 – 18 impacts, while June 10 forecasts ranged from 2 – 15. Figure 10 illustrates forecasts based on non-overlapping eight-week windows. Observed impacts generally fell within forecast ranges up to four weeks; beyond that, accuracy declined—expected given generalized lead times and varying impact accumulation periods. Plant impacts from January – July 2024 were excluded because they originated from the 2023 drought events. Forecasts relied on the XGBoost model, which outperformed others when using DSCI, ESI, and impact data. These forecasts can guide preparedness and relief programs, such as the USDA's Livestock Forage Disaster Program, which typically respond after drought onset. Forecast ranges should inform minimum and maximum relief needs.

Forecasting drought impacts is complex due to the interactions among ecosystems (plants, water, agriculture, and fire) and the cascading effects on human systems (society and relief). Impacts vary in timing: some (*e.g.*, plants) require prolonged drought, while others occur within 1–2 weeks. Monthly forecasts aggregated weekly predictions (Figure 11). Mixed lead-time forecasts were also shown, excluding plant impacts. The model predicted both reported and unreported impacts, possibly due to historical patterns or inaccuracies in reporting. Drought indices (*i.e.*, DSCI, ESI) exhibited low values from March to July 2024, while impacts increased, indicating thresholds of drought accumulation before impacts occur (Figure 12).

A comparison between observed and forecasted impacts for June 24 – July 9, 2024, with lead times of 2 – 8 weeks, is shown in Figure 13. Relief and fire impacts aligned spatially in central regions, while agricultural impacts were forecasted but not reported in several counties. This discrepancy may indicate model error or underreporting. Population distribution analysis could reveal reporting bias, as noted by Noel (2020).

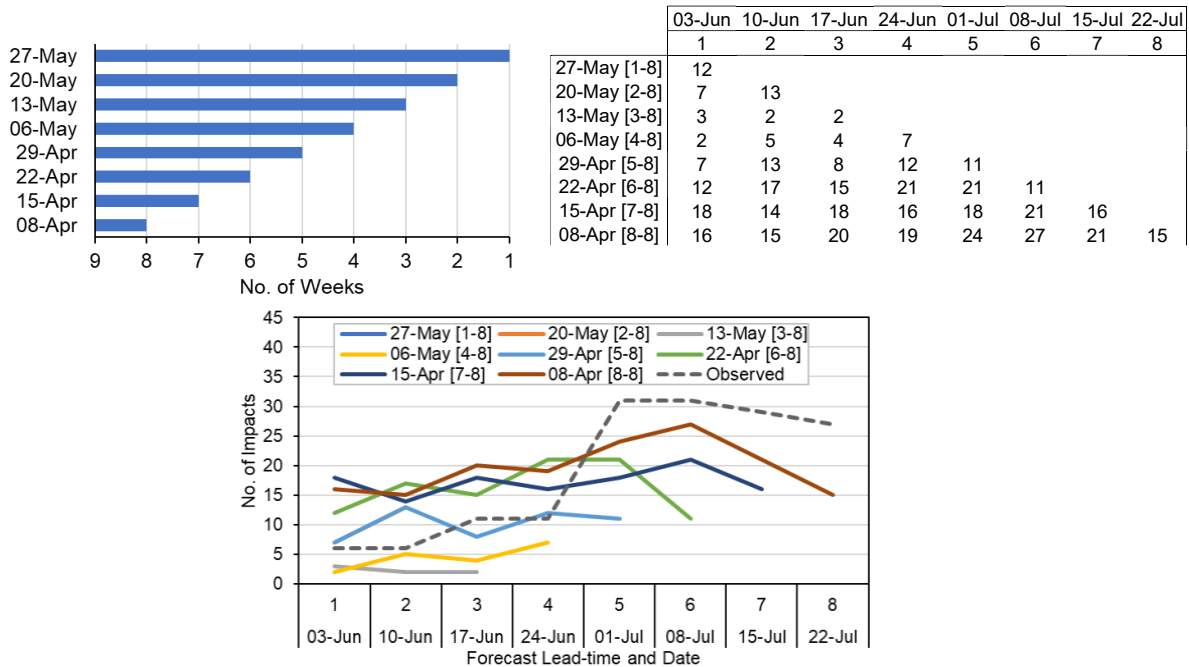

Figure 10: The top left panel shows the different periods of the data of the prior 8 weeks before June 3, 2024. For example, the data from May 20 to April 8 represents 2 – 8 past weeks of data. The top right panel shows the forecast of impacts based on the fixed, non-overlapping past 8 weeks of data, which potentially forecasts an impact on July 22. The bottom panel displays the range of all possible drought impact forecasts based on the past 8 weeks, from June 3, along with the reported impacts.

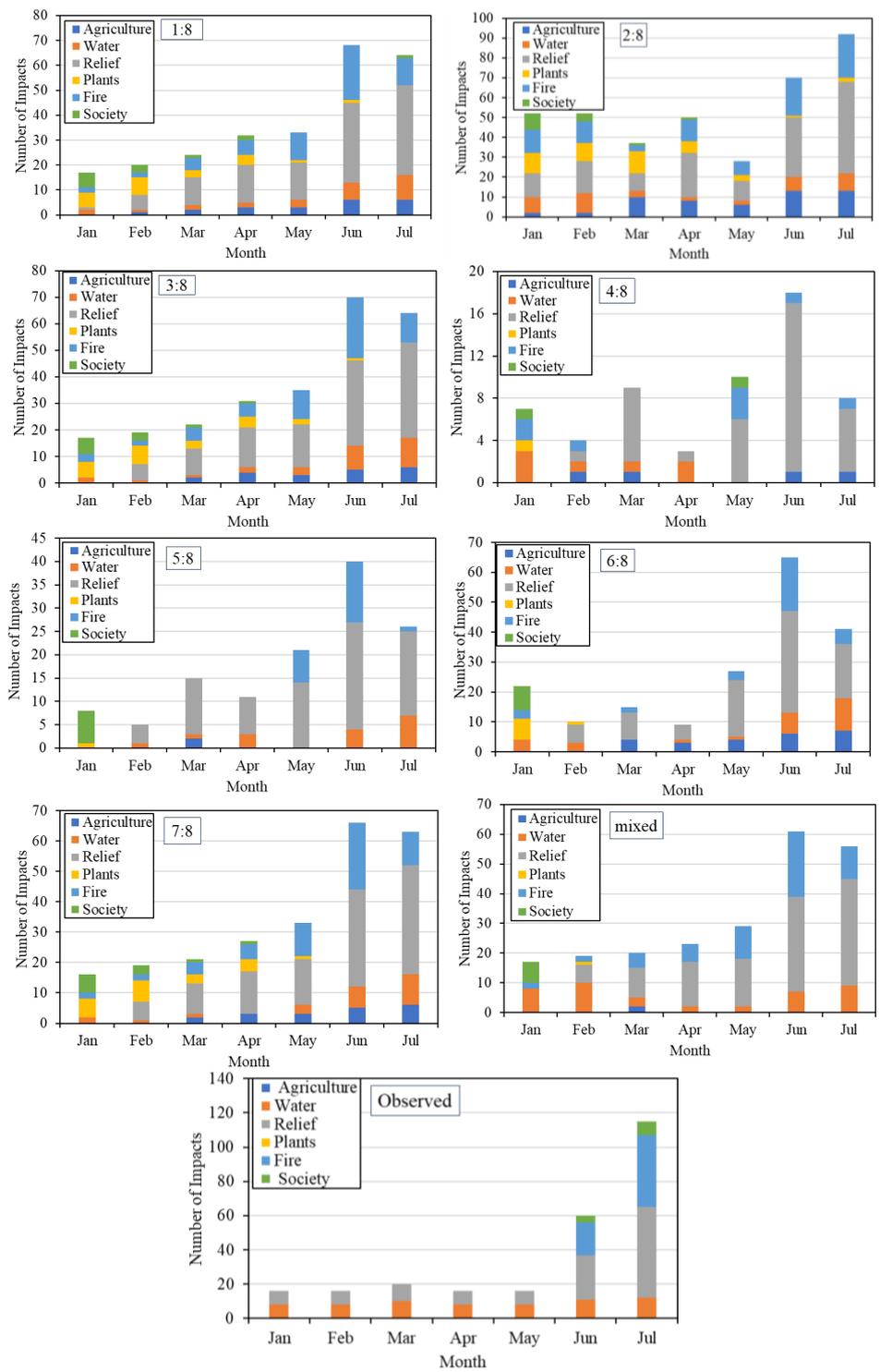

Figure 11: Monthly (weekly aggregates) of drought impact forecasts at the state level for the period between January – July 2024. The panel with 1:8 indicates predicted impacts based on data from the prior 1–8 weeks, and so forth. "Mixed" indicates predictions with variable past weeks data depending on the impact, and "Observed" refers to observed (or reported impacts, excluding Plants).

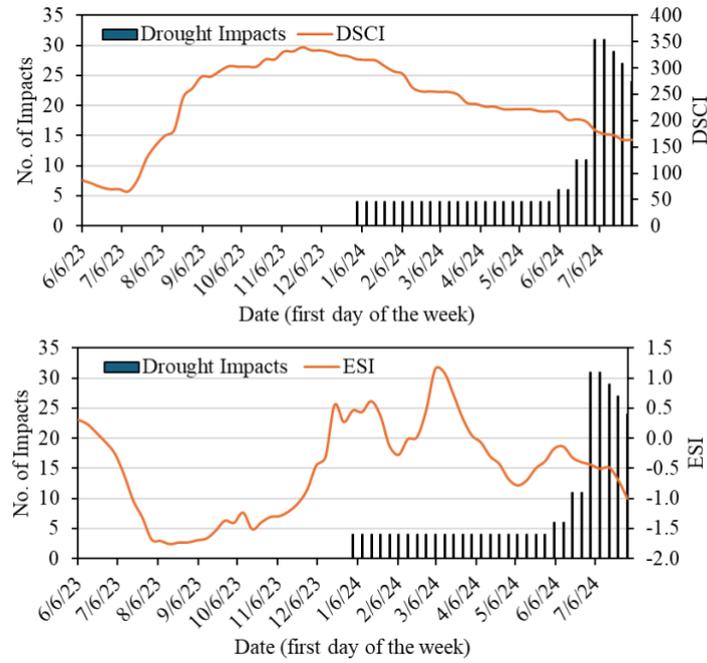

Figure 12: Plot of the Drought Severity Coverage Index (DSCI), the Evaporative Stress Index (ESI), and the reported drought impacts for the period between June 6, 2023, and July 29, 2024. It should be noted that the impacts were shown only starting from January 1, 2024.

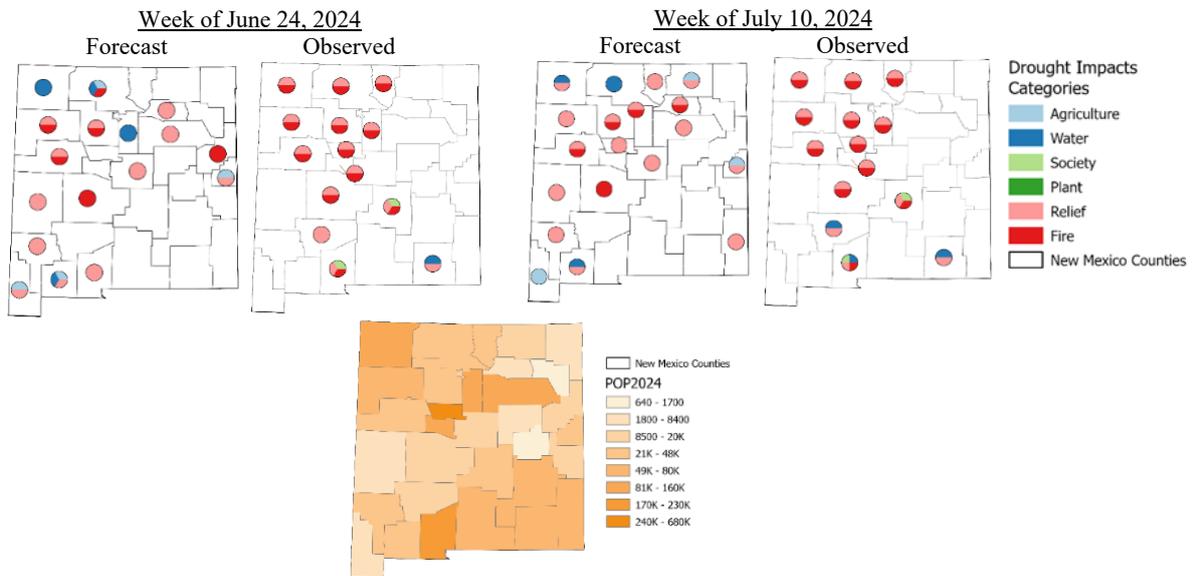

Figure 13: Drought impact forecasts for the weeks of June 24 and July 10, 2024, along with the corresponding observed (reported) impacts. The forecasts are based on XGBoost, using both DSCI and ESI, with data from the past 2–8 prior weeks.

## 6. Conclusions

The goal of this study was to predict drought impacts and forecast their occurrence with a reasonable lead time. Providing such information can help stakeholders, land managers, and decision-makers develop and implement effective drought mitigation and adaptation strategies. To accomplish this, the study addressed key conceptual challenges stemming from the complex and multidimensional nature of drought. The study utilized the Drought Severity Coverage Index (DSCI) in conjunction with the Evaporative Stress Index (ESI) to more accurately characterize drought events. The DSCI is a quasi-observational index used to depict drought conditions, while the ESI better represents rapidly intensifying droughts. The detailed text-based reports from the Drought Impact Reporter (DIR) provided a reasonable representation of historic impacts since 2005. An evaluation of three machine learning techniques—Random Forest (RF), Long Short-Term Memory (LSTM), and eXtreme Gradient Boosting (XGBoost)—showed that XGBoost most effectively captured the nonlinear relationships between drought indices and their associated impacts. The combined use of the DSCI and ESI yielded the best model predictions and forecasts. Fire and Relief impacts exhibited the highest consistency and the greatest predictive accuracy. Agriculture and Water predictions also performed well, although their accuracy began to decline more noticeably as the lead time increased. Plants and Society predictions were the most inconsistent. The final XGBoost model achieved successful forecasts up to eight weeks in advance for most impact categories. The findings of this study can help stakeholders develop and implement effective drought mitigation and adaptation strategies.

These results provide valuable guidance for stakeholders, land managers, and decision-makers seeking to better anticipate drought-related impacts and enhance preparedness strategies.

**Acknowledgments and Funding**
Funding for this research was partially provided by NOAA – National Integrated Drought Information System (NIDIS) and New Mexico State University.

**Author Contributions**
**Hatim Geli** provided the data and conceived the initial idea of using machine learning to build a model for drought impact prediction. He explained what the impacts and indices meant. He helped with the analysis of the data/results, contributed to the revision of the experimental results, and revised the paper draft. **Islam Omar** conceived and designed the experiments, built the machine learning models, analyzed and visualized the data and the results, and wrote the initial draft of the paper. **Mona Elshinawy** assisted with the machine learning models, reviewed, and contributed to the experiments and results. She extensively revised the manuscript. **Abdel-Hameed Badawy** helped with the machine learning models and contributed to the design of the experiments and the visualization of the results. He extensively revised the manuscript. **David DuBios** guided the research design and the use of drought indices. **Lara Prehodko** provided guidance on research design. **Kelly Helm Smith** provided Drought Impact Reporter data aggregated by weekly impact span, a novel format developed specifically for this project.

**Conflict of interest**
The authors declare that they have no known competing financial interests or personal relationships that could be perceived to have influenced the research findings reported in this article.

**Data Availability**
The datasets that support the findings of this study are openly available at the Drought Impact Reporter (DIR): https://unldroughtcenter.maps.arcgis.com/apps/dashboards/46afe627bb60422f85944d70069c09cf, Drought Severity and Coverage Index (DSCI): https://droughtmonitor.unl.edu/Data/DataTables.aspx, and Evaporative Stress Index (ESI): https://climateserv.servirglobal.net/map

# References

Abatzoglou J T, McEvoy D J, and Redmond K T 2017 The West Wide Drought Tracker: Drought Monitoring at Fine Spatial Scales *Bull. Amer. Meteor. Soc.* **98** 1815–20

Akyuz F A 2017 *Drought Severity and Coverage Index. United States Drought Monitor* (Lincoln, NE: The University of Nebraska-Lincoln) Online: https://droughtmonitor.unl.edu/About/AboutheData/DSCI.aspx

Anderson M C, Kustas W P, Norman J M, Hain C R, Mecikalski J R, Schultz L, González-Dugo M P, Cammalleri C, d'Urso G, Pimstein A and Gao F 2011 Mapping daily evapotranspiration at field to continental scales using geostationary and polar orbiting satellite imagery *Hydrology and Earth System Sciences* **15** 223–39

Anderson M C, Norman J M, Mecikalski J R, Otkin J A and Kustas W P 2007a A climatological study of evapotranspiration and moisture stress across the continental United States based on thermal remote sensing: 1. Model formulation *Journal of Geophysical Research: Atmospheres* **112**

Anderson M C, Norman J M, Mecikalski J R, Otkin J A and Kustas W P 2007b A climatological study of evapotranspiration and moisture stress across the continental United States based on thermal remote sensing: 2. Surface moisture climatology *Journal of Geophysical Research: Atmospheres* **112**

Bachmair S, Kohn I and Stahl K 2015 Exploring the link between drought indicators and impacts *Natural Hazards and Earth System Sciences* **15** 1381–97

Bachmair S, Svensson C, Hannaford J, Barker L and Stahl K 2016 A quantitative analysis to objectively appraise drought indicators and model drought impacts *Hydrology and Earth System Sciences* **20** 2589–609

Bachmair S, Svensson C, Prosdocimi I, Hannaford J and Stahl K 2017 Developing drought impact functions for drought risk management *Natural Hazards and Earth System Sciences* **17** 1947–60

Chen T and Guestrin C 2016 XGBoost: A Scalable Tree Boosting System Online: http://doi.acm.org/10.1145/2939672.2939785

Crausbay S D, Ramirez A R, Carter S L, Cross M S, Hall K R, Bathke D J, Betancourt J L, Colt S, Cravens A E and Dalton M S 2017 Defining ecological drought for the twenty-first century *Bulletin of the American Meteorological Society* **98** 2543–50

DIR 2025 Home | Drought Impacts Toolkit *Drought Impacts Toolkit'* Online: https://droughtimpacts.unl.edu/Home.aspx

DIR Dashboard 2025 NDMC Drought Impact Reporter Online: https://unldroughtcenter.maps.arcgis.com/apps/dashboards/46afe627bb60422f85944d70069c09cf

Gedefaw M G, Geli H M E and Abera T A 2021 Assessment of Rangeland Degradation in New Mexico Using Time Series Segmentation and Residual Trend Analysis (TSS-RESTREND) *Remote Sensing* **13** 1618

Geli H M E 2025 A Communication Framework for Ecological Drought Information in the Southwestern US | Drought.gov Online: https://www.drought.gov/drought-research/communication-framework-ecological-drought-information-southwestern-

us

Jin S, Dewitz J, Danielson P, Granneman B, Costello C, Smith K and Zhu Z 2023 National Land Cover Database 2019: A New Strategy for Creating Clean Leaf-On and Leaf-Off Landsat Composite Images *J Remote Sens* **3** 0022

Johnson L E, Geli H M E, Hayes M J and Smith K H 2020 Building an Improved Drought Climatology Using Updated Drought Tools: A New Mexico Food-Energy-Water (FEW) Systems Focus *Front. Clim.* **2** Online: https://www.frontiersin.org/journals/climate/articles/10.3389/fclim.2020.576653/full

Leeper R D, Bilotta R, Petersen B, Stiles C J, Heim R, Fuchs B, Prat O P, Palecki M and Ansari S 2022 Characterizing US drought over the past 20 years using the US drought monitor *International Journal of Climatology* **42** 6616–30

Lemaître G, Nogueira F and Aridas C K 2017 Imbalanced-learn: A Python Toolbox to Tackle the Curse of Imbalanced Datasets in Machine Learning Online: http://jmlr.org/papers/v18/16-365

NOAA NCEI 2025 NOAA National Centers for Environmental Information (NCEI) U.S. Billion-dollar Weather and Climate Disasters (2025) Online: https://www.ncei.noaa.gov/archive/accession/0209268

NOAA NIDIS 2025 Drought Early Warning | Drought.gov *l Integrated Drought Information System* Online: https://www.drought.gov/about/drought-early-warning

Noel M, Bathke D, Fuchs B, Gutzmer D, Haigh T, Hayes M, Poděbradská M, Shield C, Smith K and Svoboda M 2020 Linking Drought Impacts to Drought Severity at the State Level *Bulletin of the American Meteorological Society* Online: https://doi.org/10.1175/BAMS-D-19-0067.1

Otkin J A, Anderson M C, Hain C and Svoboda M 2015 Using temporal changes in drought indices to generate probabilistic drought intensification forecasts *J Hydrometeorol* **16** 88–105

Otkin J A, Anderson M C, Hain C, Svoboda M, Johnson D, Mueller R, Tadesse T, Wardlow B and Brown J 2016 Assessing the evolution of soil moisture and vegetation conditions during the 2012 United States flash drought *Agr Forest Meteorol* **218–219** 230–42

Otkin J A, Svoboda M, Hunt E D, Ford T W, Anderson M C, Hain C and Basara J B 2018 Flash Droughts: A Review and Assessment of the Challenges Imposed by Rapid-Onset Droughts in the United States Online: https://journals.ametsoc.org/view/journals/bams/99/5/bams-d-17-0149.1.xml

PRISM 2025 PRISM Group at Oregon State University Online: https://prism.oregonstate.edu/

Raghav P and Kumar M 2024 Are the ecosystem-level evaporative stress indices representative of evaporative stress of vegetation? *Agricultural and Forest Meteorology* **357** 110195

Redmond K T 2002 The depiction of drought: A commentary *Bulletin of the American Meteorological Society* **83** 1143–8

Sandholt I, Rasmussen K and Andersen J 2002 A simple interpretation of the surface

temperature/vegetation index space for assessment of surface moisture status *Remote Sens Environ* **79** 213–24

SERVIR GLOBAL 2025 ClimateSERV - Home Online: https://climateserv.servirglobal.net/

Smith K H, Tyre A J, Tang Z, Hayes M J and Akyuz F A 2020 Calibrating human attention as indicator: Monitoring #drought in the Twittersphere *Bulletin of the American Meteorological Society* Online: https://doi.org/10.1175/BAMS-D-19-0342.1

Sutanto S J, van der Weert M, Wanders N, Blauhut V and Van Lanen H A 2019 Moving from drought hazard to impact forecasts *Nature communications* **10** 1–7

Svoboda M, LeComte D, Hayes M, Heim R, Gleason K, Angel J and Rippey B 2002 The drought monitor *Bulletin of the American Meteorological Society* **83** 1181–90

UNDRR 2021 *GAR Special Report on Drought 2021* (United Nation Office for Disaster Risk Reduction)

USDM 2020 United States Drought Monitor - Tabular Data Archive *The United States Drought Monitor* Online: https://droughtmonitor.unl.edu/Data/DataTables.aspx

USDM 2019 What is the U.S. Drought Monitor?" United States Drought Monitor *The United States Drought Monitor* Online: https://droughtmonitor.unl.edu/AboutUSDM/WhatIsTheUSDM.aspx

USDM - DSCI 2021 Drought Severity and Coverage Index | U.S. Drought Monitor Online: https://droughtmonitor.unl.edu/About/AboutheData/DSCI.aspx

Van Loon A F, Gleeson T, Clark J, Van Dijk A I J M, Stahl K, Hannaford J, Di Baldassarre G, Teuling A J, Tallaksen L M, Uijlenhoet R, Hannah D M, Sheffield J, Svoboda M, Verbeiren B, Wagener T, Rangecroft S, Wanders N and Van Lanen H A J 2016a Drought in the Anthropocene *Nature Geosci* **9** 89–91

Van Loon A F, Stahl K, Di Baldassarre G, Clark J, Rangecroft S, Wanders N, Gleeson T, Van Dijk A I J M, Tallaksen L M and Hannaford J 2016b Drought in a human-modified world: reframing drought definitions, understanding, and analysis approaches *Hydrol Earth Syst Sc* **20** 3631

Wilhite D A 2000 Drought planning and risk assessment: Status and future directions *Annals of arid Zone* **39** 211–30

Wilhite D A and Glantz M H 1985 Understanding: the Drought Phenomenon: The Role of Definitions *Water International* **10** 111–20

Wilhite D A, Svoboda M D and Hayes M J 2007 Understanding the complex impacts of drought: A key to enhancing drought mitigation and preparedness *Water Resour Manage* **21** 763–74

WMO 2025 *Baseline Assessment of Drought Impact Monitoring* (United Nations - World Meteorological Organization) Online: https://www.un-ilibrary.org/content/books/9789263113559

WMO - GWP 2016 *Handbook of Drought Indicators and Indices* (World Meteorological Organization - Global Water Partnership) Online: https://library.wmo.int/records/item/55169-handbook-of-drought-indicators-and-indices?offset=1

WRCC 2025 West Wide Drought Tracker Online: https://wrcc.dri.edu/wwdt/time/.

WRI 2020 Aqueduct Water Risk Atlas - World Resources Institute *Aqueduct Water Risk Atlas* Online: https://www.wri.org/resources/maps/aqueduct-water-risk-atlas

Yadav K, Geli H, Cibils A F, Hayes M, Fernald A, Peach J, Sawalhah M N, Tidwell V C, Johnson L E, Zaied A J and Gedefaw M G 2021 An Integrated Food, Energy, and Water Nexus, Human Wellbeing, and Resilience (FEW-WISE) Framework: New Mexico *Front. Environ. Sci.* **9** Online: https://www.frontiersin.org/articles/10.3389/fenvs.2021.667018/abstract

Zhang B, Salem F K A, Hayes M J, Smith K H, Tadesse T and Wardlow B D 2023 Explainable machine learning for the prediction and assessment of complex drought impacts *Science of The Total Environment* **898** 165509